\documentclass[10pt,journal,compsoc]{IEEEtran}
\makeatletter
\def\endthebibliography{%
	\def\@noitemerr{\@latex@warning{Empty `thebibliography' environment}}%
	\endlist
}

\makeatother

\ifCLASSOPTIONcompsoc
  \usepackage[nocompress]{cite}
\else
  \usepackage{cite}
\fi
\usepackage[vlined,ruled,linesnumbered]{algorithm2e}
\usepackage{stfloats}
\usepackage{epsfig}
\usepackage{graphicx}
\usepackage{amsmath}
\usepackage{amssymb}
\usepackage{multirow}
\usepackage{url}
\usepackage{threeparttable}
\usepackage{booktabs}
\usepackage{dcolumn}
\usepackage{graphicx}
\usepackage{float} 
\usepackage{subfig}
\usepackage{enumitem}
\usepackage{float}
\newcolumntype{d}[1]{D{.}{.}{#1}}

\usepackage[pagebackref=false,breaklinks=true,letterpaper=true,colorlinks,bookmarks=false,citecolor={blue},linkcolor={blue}]{hyperref}

\usepackage{caption}
\usepackage[margin=4pt,font=footnotesize,labelfont=bf,labelsep=endash,tableposition=top]{caption}

\begin{document}

\title{Structured Knowledge Distillation for\\ Dense Prediction}

\author{Yifan~Liu, %
        Changyong~Shu, %
        Jingdong Wang, %
        Chunhua Shen%
\IEEEcompsocitemizethanks
{\IEEEcompsocthanksitem Y. Liu and C. Shen are with 
The University of Adelaide, SA 5005, Australia. 
\protect\\
E-mail: \{yifan.liu04, chunhua.shen\}@adelaide.edu.au
\IEEEcompsocthanksitem
C. Shu is with Nanjing Institute of Advanced Artificial Intelligence, China.
\IEEEcompsocthanksitem J. Wang is with Microsoft Research, Beijing, China.\protect \\
}
}%

\IEEEtitleabstractindextext{%
\begin{abstract}

In this work,
we consider transferring the structure information from large networks to 
compact 
ones 
for dense prediction 
tasks in computer vision. 
Previous knowledge distillation strategies 
used 
for dense prediction 
tasks
often %
directly borrow
the distillation scheme  %
for image classification %
and perform knowledge distillation for each pixel~\emph{separately},
leading to sub-optimal performance. 
Here we propose to distill 
\emph{structured} knowledge
from large networks to compact  networks, 
taking into account  the fact  
that dense prediction is a structured prediction problem.
Specifically, we study two structured distillation schemes:
\textit{i}) \emph{pair-wise} distillation
that distills the pair-wise similarities by building a static graph;
and 
\textit{ii}) \emph{holistic} distillation that uses
adversarial training 
to distill holistic knowledge.
The effectiveness of our knowledge distillation approaches
is demonstrated 
by
experiments on three dense prediction tasks: semantic segmentation, depth estimation and object detection.
Code is available at:
\url{https://git.io/StructKD}

\end{abstract}

\begin{IEEEkeywords}
Structured  knowledge distillation, adversarial training, knowledge transferring, dense prediction.
\end{IEEEkeywords}}

\maketitle

\IEEEdisplaynontitleabstractindextext

\IEEEpeerreviewmaketitle

\section{Introduction}
Dense prediction is a 
family 
of fundamental problems in computer vision, which learns a mapping from input %
images 
to complex output structures, %
including 
semantic segmentation, depth estimation and object detection, among many others.
One needs to assign category labels or regress specific values for each pixel given an input image to form the structured outputs.
In general these tasks are significantly more challenging to solve than image-level prediction problems, thus often requiring networks with large capacity in order to achieve satisfactory accuracy. On the other hand,  compact models are desirable for enabling computing  
on 
edge devices  with limited computation resources.

Deep neural networks have been the dominant solutions since the invention of fully-convolutional neural networks (FCNs)~\cite{shelhamer2017fully}. Subsequent approaches, e.g., DeepLab~\cite{%
chen2018deeplab%
}, PSPNet~\cite{zhao2017pyramid}, RefineNet~\cite{lin2019refinenet}, 
and FCOS~\cite{tian2019fcos} follow
the design of FCNs to optimize energy-based objective functions related to different tasks, having achieved significant improvement in accuracy, often with cumbersome models and expensive computation.

Recently, design of neural networks 
with compact model sizes, light computation cost
and high performance, has attracted much attention
due to 
the need of applications on mobile devices.
Most current efforts have been devoted to
designing lightweight networks
specifically 
for dense prediction tasks
or borrowing the design from classification networks,
e.g., ENet \cite{paszke2016enet}, ESPNet \cite{paszke2016enet} and
ICNet \cite{zhao2017icnet} for semantic segmentation, YOLO \cite{redmon2016you} and
SSD \cite{liu2016ssd} for object detection, and FastDepth~\cite{wofk2019fastdepth} for depth estimation.
Strategies such as 
pruning~\cite{wofk2019fastdepth}, knowledge distillation~\cite{Li2017MimickingVE,xieimproving} are applied to 
helping
the training of compact networks by making use of cumbersome networks.

\begin{figure*}[ht]
\centering  %
\includegraphics[width=0.384028\textwidth]{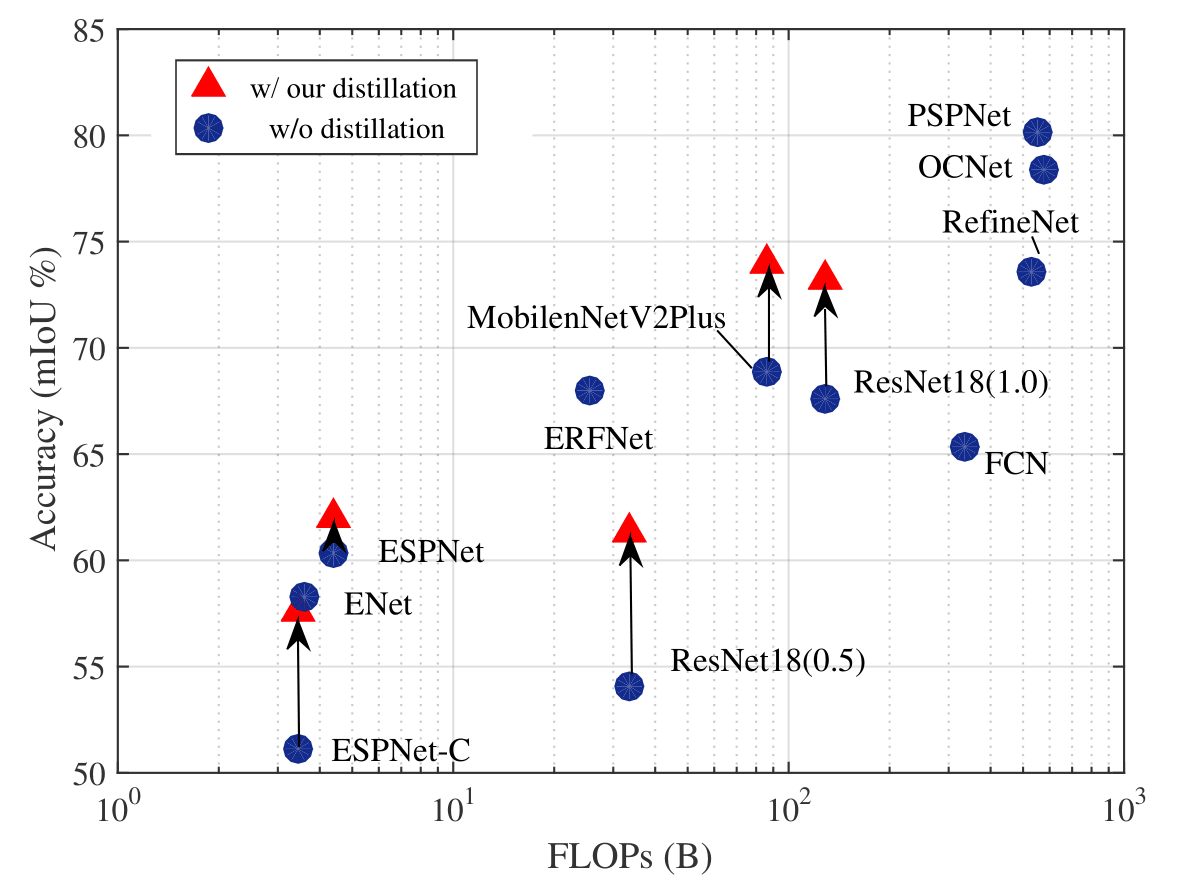}
\includegraphics[width=0.384028\textwidth]{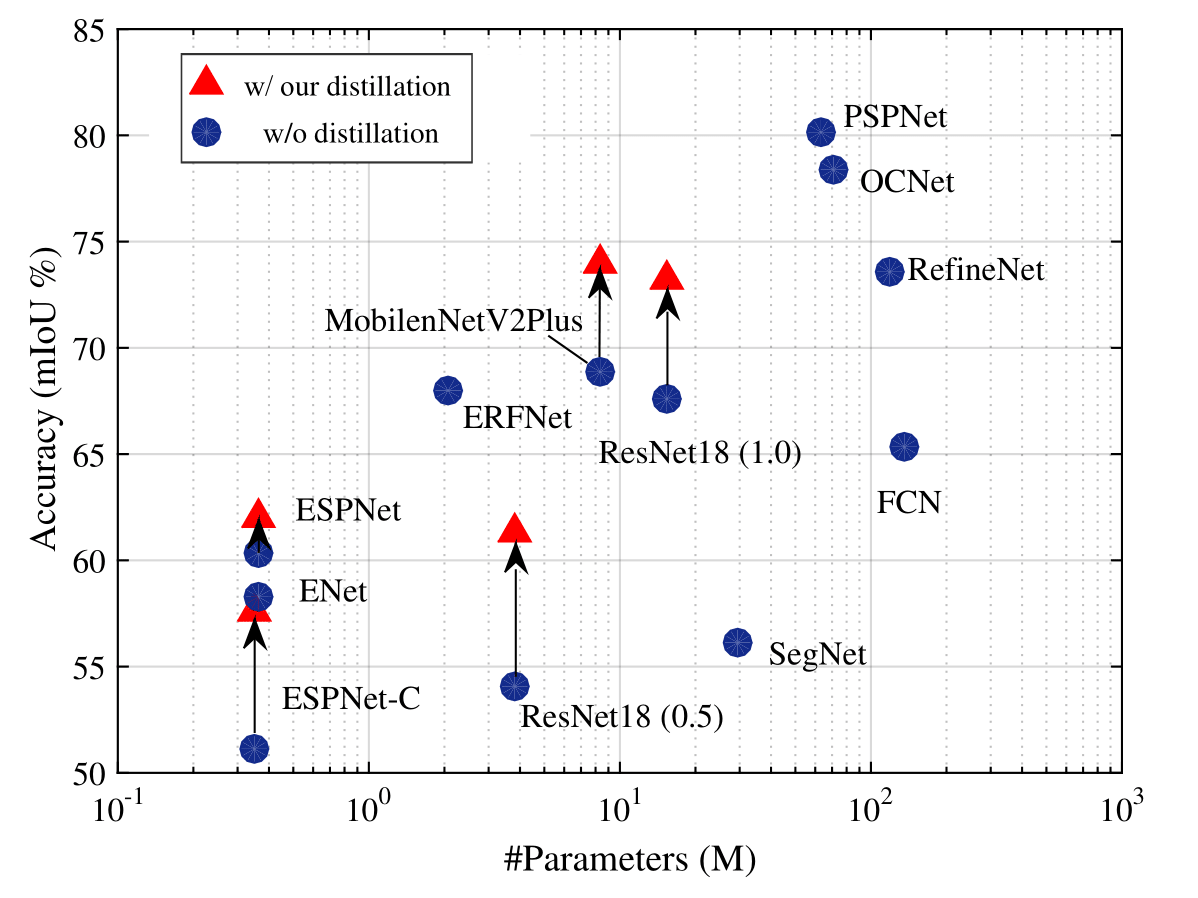}
\caption{An example on 
the
semantic segmentation task shows comparisons in terms of computation complexity, number of parameters and %
mIoU for different networks on the Cityscapes test set. The FLOPs is calculated with the input resolution of $512 \times 1024$ pixels. Red triangles are the results of our distillation method while others are without distillation. Blue circles are collected from FCN* \cite{shelhamer2017fully},  RefineNet \cite{lin2019refinenet}, SegNet \cite{badrinarayanan2017segnet}, ENet \cite{paszke2016enet}, PSPNet \cite{zhao2017pyramid}, ERFNet \cite{romera2017efficient}, ESPNet \cite{mehta2018espnet}, MobileNetV2Plus \cite{LightNet}, and OCNet \cite{Yuan2018OCNetOC}. %
With our proposed distillation method, we can achieve a higher mIoU, with no extra FLOPs and parameters.}
\label{Fig.iou_acc}
\end{figure*}
Knowledge distillation 
has 
proven effective 
in  
training compact models for 
classification tasks \cite{hinton2015distilling,romero2014fitnets}. 
As for dense prediction tasks, 
most 
previous works~\cite{Li2017MimickingVE,xieimproving} directly apply distillation %
at 
each pixel separately to transfer the class probability
from the cumbersome network (teacher) to the compact network (student);
or to 
extract 
more discriminative feature embeddings for the compact network.
Note that,  
such a pixel-wise distillation scheme neglects the important structure information. 

Considering the characteristic of dense prediction problem,
here 
we present structured knowledge distillation and transfer the structure information with two schemes,
\emph{pair-wise distillation}
and \emph{holistic distillation}.
The \emph{pair-wise distillation} scheme
is motivated by
the widely-studied 
pair-wise Markov random field framework \cite{li2009markov} for
enforcing
spatial labeling 
consistency. 
The goal is 
to align a static affinity graph which is computed to capture both short and long range structure information among different locations from the compact network
and the 
teacher 
network.

The \emph{holistic distillation} scheme
aims to align higher-order consistencies,
which are not characterized 
in the pixel-wise and pair-wise distillation,
between output structures produced from the compact network and the 
teacher 
network. 
We adopt the adversarial
training scheme, and a fully convolutional network, a.k.a.\ 
the discriminator, 
considering 
both the input image and the output structures to produce a holistic embedding which represents the quality of the structure. The compact network is encouraged to generate structures with similar embeddings as the %
teacher network. We 
encode 
the 
structure knowledge 
into the weights of discriminators.

To this end, 
we optimize an objective function
that combines a conventional task loss with
the distillation terms.
The main contributions of this paper 
are 
as follows.
\begin{itemize}
\item We study the knowledge distillation strategy for training accurate compact  networks for dense prediction.
\item We present two structured knowledge distillation schemes, pair-wise distillation and holistic distillation, enforcing pair-wise and high-order consistency between the outputs of the compact and teacher networks.
\item We demonstrate the effectiveness of our approach by improving recent
state-of-the-art compact networks
on 
three different dense prediction tasks: semantic segmentation, depth estimation and object detection. 
Taking semantic segmentation as an example, the performance  gain is illustrated in Figure ~\ref{Fig.iou_acc}.

\end{itemize}

\subsection{Related Work}
\noindent\textbf{Semantic segmentation.}
Semantic segmentation is a 
pixel classification problem, which requires an
semantic 
understanding of the whole scene. 
Deep convolutional neural networks
have been the dominant solution
to semantic segmentation
since the pioneering work, 
fully-convolutional network~\cite{shelhamer2017fully}.
Various schemes
have been developed
for improving the network capability and 
accordingly the segmentation performance.
For example, stronger backbone networks
such as 
ResNet \cite{He2016DeepRL}, DenseNet \cite{huang2017densely} and HRNet~\cite{SunXLW19,hrnet},
have shown %
improved 
segmentation performance.
Retaining 
the spatial resolution 
through dilated convolutions~\cite{%
chen2018deeplab%
} or multi-path refine networks~\cite{lin2019refinenet}
leads to significant performance gain.
Exploiting multi-scale context %
using 
dilated convolutions~\cite{yu2015multi}, or pyramid pooling modules 
as 
in PSPNet~\cite{zhao2017pyramid},
also benefits the segmentation.
Lin et al.\  %
\cite{lin2016efficient}
combine deep models with structured output learning 
for semantic segmentation.

Recently, highly efficient segmentation networks have been 
attracting increasingly more interests
due to the need of %
real-time 
and  
mobile applications.
Most works focus on lightweight network design
by accelerating the convolution operations
with techniques such as  factorization techniques.
ENet~\cite{paszke2016enet}, inspired by~\cite{SzegedyVISW16}, integrates 
several acceleration factors,
including multi-branch modules,
early feature map resolution down-sampling,
small decoder size,
filter tensor factorization,
and so on.
SQ~\cite{treml2016speeding} adopts 
the SqueezeNet~\cite{IandolaMAHDK16} fire modules and parallel dilated convolution layers for efficient segmentation.
ESPNet \cite{mehta2018espnet} proposes an efficient spatial pyramid,
which is based on filter factorization techniques: point-wise convolutions and spatial pyramid of dilated convolutions,
to replace the standard convolution.
The efficient classification networks such as  
MobileNet~\cite{howard2017mobilenets} and ShuffleNet~\cite{Zhang2017ShuffleNet}
are also applied to accelerate segmentation.
In addition,
ICNet (image cascade network)~\cite{zhao2017icnet} exploits the efficiency of processing
low-resolution images and high inference quality of high-resolution ones, 
achieving a trade-off between efficiency and accuracy.

\noindent\textbf{Depth estimation.}
Depth estimation from a monocular image is essentially an ill-posed problem, which requires an expressive model with
large capacity. 
Previous works
depend on hand-crafted features \cite{saxena2007learning}.
Since 
Eigen et al.\  \cite{eigen} propose to use deep learning to 
predict %
depth maps, following works \cite{CVPR15Liu,Depth2015Liu,li2018deep,fu2018deep} benefit from the increasing 
capacity 
of %
deep models 
and achieve 
good 
results. Besides, Fei~\cite{fei2018geo} propose a semantically informed geometric loss while Yin et al.~\cite{wei2019enforcing} 
introduce 
a virtual normal loss to
exploit 
geometric 
information.
As 
in semantic segmentation, 
some works try to replace the encoder with 
efficient 
backbones~\cite{wei2019enforcing,spek2018cream,wofk2019fastdepth} to decrease the computational cost, but often suffer from 
the limitation caused by the capacity of a compact network.
The 
close 
work to ours \cite{wofk2019fastdepth} applies pruning 
to 
train 
the compact depth network. 
However,
here  we focus on the structured knowledge distillation.

\noindent\textbf{Object detection.}
Object detection is a fundamental task in computer vision, in which one needs to regress
the location of 
bounding boxes  
and 
predict a category label for each instance of interest in an image.  %
Early works \cite{girshick2015fast,ren2015faster}
achieve good 
performance by first predicting proposals and then refining the localization of bounding boxes. 
Effort is also spent on improving 
detection efficiency 
with methods 
such as  Yolo~\cite{redmon2016you}, and SSD~\cite{liu2016ssd}. 
They use a one-stage method and design light-weight network structures. 
RetinaNet~\cite{lin2017focal} solves the problem of unbalance samples %
to some extent by proposing the focal loss, which makes the results of one-stage methods comparable to two-stage ones. 
Most of the above detectors rely on a set of pre-defined anchor boxes, which decreases the training samples and makes the detection network sensitive to hyper parameters. Recently, anchor free methods 
show promises, e.g., 
FCOS~\cite{tian2019fcos}. 
FCOS employs a fully convolutional framework, and predicts bounding box based on every pixels 
same as 
in semantic segmentation, which solves the object detection task as a dense prediction problem. In this work, we apply the structured knowledge distillation method with the FCOS framework, as it is simple and 
achieves 
promising 
performance.

\noindent\textbf{Knowledge distillation}.
Knowledge distillation~\cite{hinton2015distilling}
transfers 
knowledge from
a cumbersome
model  to a compact model 
so as to improve the performance of compact networks.
It has been applied to image classification
by using the class probabilities produced from the cumbersome model as ``soft targets'' for training the
compact model~\cite{ba2014deep, hinton2015distilling, Urban2016Do}
or transferring the intermediate feature maps~\cite{romero2014fitnets, Zagoruyko2016PayingMA}.

The MIMIC
\cite{Li2017MimickingVE} method distills a compact object detection network by making use of a two-stage Faster-RCNN~\cite{ren2015faster}. They align the feature map 
at 
the 
pixel level and %
do not make use of 
the structure information among pixels.

The %
recent and independent
work,  
which applies distillation to 
semantic segmentation~\cite{xieimproving}, 
is related to our approach. 
It mainly distills the class probabilities for each pixel separately 
(as our pixel-wise distillation)
and center-surrounding differences of labels for each local patch 
(termed as a local relation in~\cite{xieimproving}). In contrast,
we focus on distilling structured knowledge: pair-wise distillation,
which transfers the relation among different locations by building a static affinity graph, and holistic distillation,
which transfers the holistic knowledge 
that captures high-order information.
Thus the work of \cite{xieimproving} may be seen as a special case of the pair-wise distillation.

This work here 
is a substantial extension of our previous %
work appeared in 
\cite{liu2019structured}. The main difference compared with ~\cite{liu2019structured} 
 is
threefold. 1) We extend the pair-wise distillation to a more general case in Section~\ref{sec:pa} and build a static graph with nodes and connections. We explore the influence of the graph size, and find out that it is important to keep a fully connected 
graph. 2) We provide more explanations and ablations on the adversarial training for holistic distillation. 3) We also extend our method to depth estimation and object detection with  
two recent %
strong baselines %
\cite{wei2019enforcing} and %
\cite{tian2019fcos}, by replacing the backbone with MobileNetV2~\cite{Sandler2018MobileNetV2IR}, and further improve their performance.

\noindent\textbf{Adversarial learning}.
Generative adversarial networks (GANs) have been widely studied in
text generation~\cite{wang2018text,yu2017seqgan} and image synthesis~\cite{Goodfellow2014Generative,karras2017progressive}.
The conditional version~\cite{MirzaO14} 
is successfully applied to image-to-image translation,
such as 
style transfer~\cite{Johnson2016Perceptual} and image coloring~\cite{Yifan2018Auto}.

The idea of adversarial learning is also %
employed 
in pose estimation~\cite{chen2017adversarial},
encouraging the human pose estimation result
not to be distinguished from the ground-truth; and
semantic segmentation~\cite{luc2016semantic},
encouraging the estimated segmentation map
not to be distinguished from the ground-truth map.
One challenge in \cite{luc2016semantic} is the mismatch between the generator's continuous output and the discrete true labels, making the discriminator in GAN be of very limited success. Different from \cite{luc2016semantic}, in our approach, the employed GAN does not %
face 
this %
issue 
as the  ground truth for the discriminator is the teacher network's logits, which are real valued. We use adversarial learning to encourage
the alignment between the output maps
produced from the cumbersome network and the compact network. However, in the depth prediction task, the ground truth maps are not discrete labels. %
In~\cite{gwn2018generative}, the authors use the ground truth maps as the real samples. Different from theirs, 
our distillation methods 
attempt 
to align the output of the cumbersome network and that of the compact network.  The task loss calculated with ground truth is optional. When the labelled data is limited, given a well-trained teacher, our method can be applied to unlabelled data and may further improve the accuracy.

\renewcommand{\arraystretch}{1.15}
\begin{figure}[tb!]
\setlength{\belowcaptionskip}{-0.2cm}
\begin{center}
\includegraphics[width=0.45\textwidth]{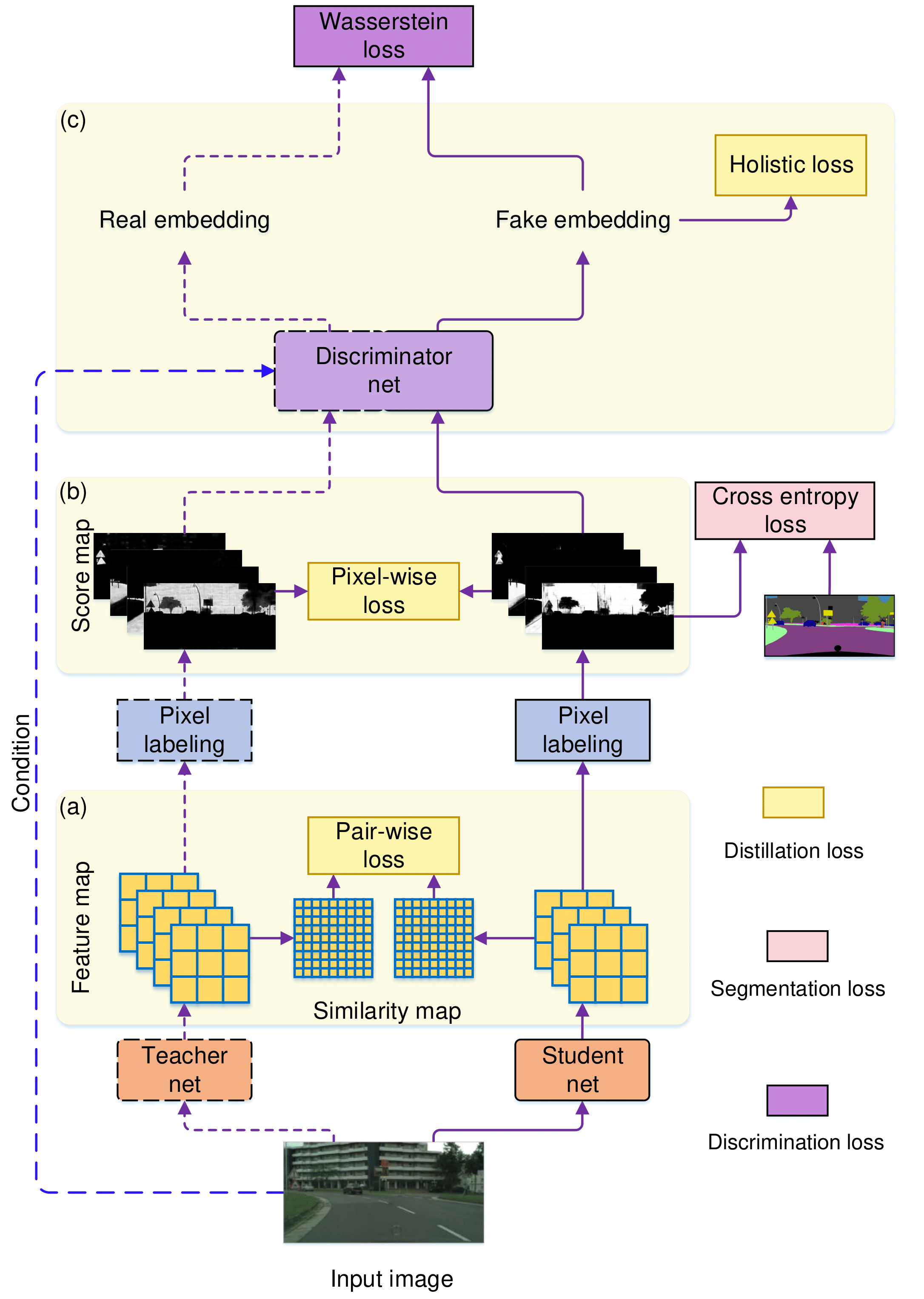}
\end{center}
\caption{Our distillation framework with the semantic segmentation task as an example. 
(a) Pair-wise distillation;
(b) Pixel-wise distillation;
(c) Holistic distillation. 
In the training process, we keep the cumbersome network  fixed as our teacher net, and only the student net and the discriminator net 
are 
optimized. The student net with a compact architecture 
 is 
trained with the three distillation terms and a task-specific loss, e.g., the cross entropy loss for  semantic segmentation.}
\label{fig:pipeline}
\end{figure}

\section{Approach}
In this section, we first introduce the structured knowledge distillation method 
for 
semantic segmentation, a task
of %
assigning 
a category label 
to each pixel in the image
from $C$ categories.
A segmentation network takes
a RGB image $\mathbf{I}$ of size $W \times H \times 3$
as the input; then it computes a feature map $\mathbf{F}$ of size $W' \times H' \times N$,
where $N$ is the number of channels.
Then, a classifier is applied to compute the segmentation map
$\mathbf{Q}$ of size $W' \times H' \times C$ from $\mathbf{F}$,
which is upsampled to the spatial size $W \times H$
of the input image to obtain the segmentation results.
We %
extend 
our method to other two dense prediction tasks: depth estimation and object detection.

\noindent\textbf{Pixel-wise distillation}.
We apply the knowledge distillation  strategy \cite{hinton2015distilling}
to 
transfer
the knowledge of the %
large teacher 
segmentation network $\mathsf{T}$
to a compact segmentation network $\mathsf{S}$
for better training the compact segmentation network.
We view  the segmentation problem
as a collection of separate pixel labeling problems,
and directly use knowledge distillation to
align the class probability of each pixel produced from the compact network.
We 
follow \cite{hinton2015distilling} and 
use the class probabilities produced from the %
teacher 
model as {soft targets} for training the compact network. 

The loss function is
as follows,
\begin{align}
\ell_{pi} (\mathsf{S}) = \frac{1}{W' \times H'} \sum_{i \in \mathcal{R}} \operatorname{KL}(\mathbf{q}^s_i \| \mathbf{q}^t_i),
\end{align}
where $\mathbf{q}^s_i$ represents  the class probabilities
of the $i$th pixel produced from the compact network $\mathsf{S}$.
$\mathbf{q}^t_i$ represents the class probabilities of the $i$th pixel produced
from the cumbersome network $\mathsf{T}$.
$\operatorname{KL}(\cdot)$ is the Kullback$\textrm{-}$Leibler divergence
between two probabilities,
and $\mathcal{R} = \{1, 2, \dots, W' \times H'\}$
denotes all the pixels.

\subsection{Structured Knowledge Distillation}

In addition to above straightforward  pixel-wise distillation,
we present two structured knowledge distillation schemes---%
pair-wise distillation and holistic distillation---%
to transfer structured knowledge from the 
teacher 
network
to the compact network.
The pipeline is illustrated in
Figure~\ref{fig:pipeline}.

\noindent\textbf{Pair-wise distillation}.
\label{sec:pa}
Inspired by the pair-wise Markov random field framework
that is widely adopted for improving spatial labeling contiguity,
we propose to transfer the pair-wise relations, 
specifically 
pair-wise similarities in our approach, among spatial locations.

\begin{figure}[b]
\setlength{\belowcaptionskip}{-0.2cm}
\centering  
\subfloat[ $\alpha=9$, $\beta=1$]{
\includegraphics[width=0.13\textwidth]{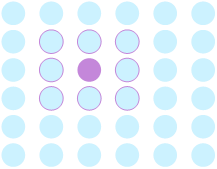}}
\subfloat[$\alpha=25$, $\beta=1$]{
\includegraphics[width=0.13\textwidth]{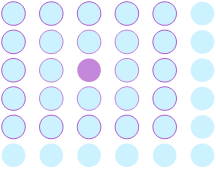}}
\subfloat[$\alpha=9$, $\beta=4$]{
\includegraphics[width=0.13\textwidth]{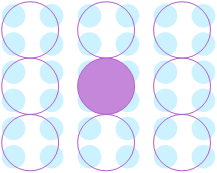}}

\caption{Illustrations of the connection range $\alpha$ and the granularity $\beta$ of each node.}

\label{fig:graph}
\end{figure}

We build 
an
affinity graph to denote the spatial pair-wise relations, in which, the nodes represent 
different spatial locations and the connection between two nodes represents the similarity.
We denote the connection range $\alpha$ and the granularity $\beta$ of each node to control the size of the static affinity graph. For each node, we only consider similarities with top-$\alpha$ near nodes according to spatial distance (here we use the Chebyshev distance) and aggregate $\beta$ pixels in a spatial local patch to represent the feature of this node as illustrate in Figure~\ref{fig:graph}. Here 
for a $W' \times H' \times C$ feature map, %
$W' \times H'$ is the spatial resolution. 
With the granularity $\beta$ and the connection range $\alpha$, the %
affinity graph  contains $\frac{W' \times H'}{\beta}$ nodes with $\frac{W' \times H'}{\beta} \times \alpha$ connections.

Let $a^t_{ij}$ and $a^s_{ij}$ denote the similarity between the $i$th node
and the $j$th node produced from the
teacher 
network $\mathsf{T}$ and the compact network $\mathsf{S}$, respectively.
We adopt the squared difference
to formulate the pair-wise similarity distillation loss,
\begin{align}
\ell_{pa} (\mathsf{S}) = \frac{\beta}{W' \times H'\times \alpha} 
\sum_{i \in \mathcal{R'}} \sum_{j \in \alpha}  
(a^s_{ij} - a^t_{ij})^2.
\end{align}
where $\mathcal{R'} = \{1, 2, \dots,\frac{W' \times H'}{\beta}\}$ denotes all the nodes.
In our implementation, we use 
average 
pooling 
to aggregate $\beta \times C$ features in one node to be $1 \times C$, and
the similarity between two nodes is computed
from the aggregated features $\mathbf{f}_i$
and $\mathbf{f}_j$
as $$a_{ij} = {\mathbf{f}_i^\top \mathbf{f}_j}/{(\|\mathbf{f}_i\|_2\|\mathbf{f}_j\|_2)},$$
which empirically works well.

\noindent\textbf{Holistic distillation}. 
\label{sec:ho}
We align high-order relations 
between the segmentation maps
produced from the teacher and compact networks.
The holistic embeddings of the segmentation maps are computed as the representations. 

We 
use 
conditional generative adversarial learning~\cite{MirzaO14}
for formulating the holistic distillation problem.
The compact net is 
viewed 
as a generator conditioned on the input RGB image $\mathbf{I}$, 
and the predicted segmentation map $\mathbf{Q}^s$ is 
seen
as a fake sample. 
The segmentation map predicted by the teacher ( $\mathbf{Q}^t$) is 
the real sample.
We expect that $\mathbf{Q}^s$ is similar to $\mathbf{Q}^t$.
Generative adversarial networks (GANs) usually suffer from the unstable gradient in training the generator due to the 
non-smoothness 
of the discontinuous 
Kullback–Leibler
(KL) or {Jensen-Shannon} (JS) divergence 
when the two distributions do not overlap. 
The Wasserstein distance~\cite{gulrajani2017improved} provides a smooth 
measure of the difference between two distributions. The Wasserstein distance is defined as the minimum cost to converge the model distribution $p_{s}(\mathbf{Q}^s)$ to the real distribution $p_{t}(\mathbf{Q}^t)$. 
This can be  written as 
follows:
\begin{align}
\ell_{ho} (\mathsf{S}, \mathsf{D}) = &~\mathbb{E}_{\mathbf{Q}^s \sim p_{s}(\mathbf{Q}^s)}[\mathsf{D}(\mathbf{Q}^s|\mathbf{I})] 
- \mathbb{E}_{\mathbf{Q}^t \sim p_{t}(\mathbf{Q}^t)}[\mathsf{D}(\mathbf{Q}^t|\mathbf{I})],
\label{eqn:holisticterm}
\end{align}
where $\mathbb{E}[\cdot]$ is the expectation operator, 
and $\mathsf{D}( \cdot )$ is an embedding network, 
acting as the discriminator in GAN,
which projects $\mathbf{Q}$ and $\mathbf{I}$ together
into a holistic embedding score.
The Lipschitz constraint is enforced by applying a gradient penalty 
\cite{gulrajani2017improved}.

The segmentation map and the RGB image are concatenated as the input of the embedding network $\mathsf{D}$. $\mathsf{D}$ is a fully convolutional neural network with five convolution blocks. Each convolution block has ReLU and BN layers, except for the final output convolution block. 
Two self-attention modules are inserted between the final three blocks to capture the structure information~\cite{Zhang2018Self}. 
We insert a batch normalization immediately after the input so as 
to normalize the scale difference between the segmentation map and RGB channels.

Such a discriminator is able to produce a holistic embedding representing how well the input image and the segmentation map match. We further add a pooling layer to pool the holistic embedding into a score. As we employ the Wasserstein distance in the adversarial training, the discriminator is trained to 
produce 
a higher score w.r.t.\ 
the output segmentation map from the teacher net and 
produce 
lower scores w.r.t.\ 
the ones from the student. In this 
process, 
we 
encode 
the knowledge of evaluating the quality of a segmentation map into the discriminator. The student is trained with the regularization of achieving a higher score under the evaluation of the discriminator,
thus
improving the performance of the student.

\subsection{Optimization}
The
overall 
objective function consists of 
a 
standard 
multi-class cross-entropy loss $\ell_{mc} (\mathsf{S})$
with pixel-wise and structured distillation terms%
\footnote{The objective function 
is the summation
of the losses over the mini-batch of training samples.
For 
ease of exposition, here
we 
omit 
the summation operator.}
\begin{align}
\ell (\mathsf{S}, \mathsf{D}) 
= &~\ell_{mc} (\mathsf{S}) + \lambda_1 (\ell_{pi}(\mathsf{S}) 
+ \ell_{pa}(\mathsf{S}))  
- \lambda_2 \ell_{ho}(\mathsf{S}, \mathsf{D}),
\label{eqn:all}
\end{align}
where $\lambda_1$ and $\lambda_2$
are set 
to 
$10$ and $0.1$,
making these loss value ranges comparable. 
We minimize the objective function with respect to the parameters of the compact segmentation
network $\mathsf{S}$, 
while maximizing  it 
w.r.t.\ 
the parameters of the discriminator $\mathsf{D}$,
which is implemented by iterating the following two steps:
\begin{itemize}
    \item \textbf{Train the discriminator $\mathsf{D}$}.
Training the discriminator 
is equivalent to minimizing $\ell_{ho}(\mathsf{S}, \mathsf{D})$. $\mathsf{D}$ aims to give a high embedding score for the real samples from the teacher net and low embedding scores for the fake samples from the student net.
    \item \textbf{Train the compact segmentation network $\mathsf{S}$}.
    Given the discriminator network,
    the goal is to minimize the multi-class cross-entropy loss and the distillation losses
    relevant to the compact segmentation network:
    $$\ell_{mc} (\mathsf{S}) + \lambda_1 (\ell_{pi}(\mathsf{S}) 
+ \ell_{pa}(\mathsf{S}))  
- \lambda_2 \ell^s_{ho}(\mathsf{S}),$$
where $$\ell^s_{ho}(\mathsf{S}) = \mathbb{E}_{\mathbf{Q}^s \sim p_{s}(\mathbf{Q}^s)}[\mathsf{D}(\mathbf{Q}^s|\mathbf{I})]$$
is a part of $\ell_{ho}(\mathsf{S}, \mathsf{D})$ given in Equation~\eqref{eqn:holisticterm}, and we expect $\mathsf{S}$ to achieve a higher score under the evaluation of $\mathsf{D}$.
\end{itemize}

\subsection{Extension to Other Dense Prediction Tasks}

Dense prediction learns a mapping from an input RGB image $\mathbf{I}$ %
of 
size $W \times H \times 3$ to a per-pixel output $\mathbf{Q}$ 
of 
size $W \times H \times C$. 
In semantic segmentation, the output has $C$ channels which is the number of semantic classes.

For the object detection task, for each pixel, we predict the $c^*$ classes, as well as a 4D vector 
$t^*=(l,t,r,b)$ representing the location of a  bounding box. We follow FCOS~\cite{tian2019fcos}, and combine them with distillation terms as regularization.

The depth estimation task can be solved as a classification task, as the continuous depth values can be divided into $C$ discrete categories~\cite{cao2017estimating}. For inference, 
we apply a soft weighted sum as in \cite{wei2019enforcing}. The pair-wise distillation can be directly applied to the intermediate feature maps. The holistic distillation uses the depth map as input. We can use the ground truth as the real samples of GAN in 
the 
depth estimation task, because it is a continuous map. However, in order to apply our method to unlabelled data, we still use 
depth maps from the teacher as our real samples.

\section{Experiments}
In this section, we first 
apply our method to 
semantic segmentation 
to 
empirically 
verify the effectiveness of structured knowledge distillation. We discuss and explore how 
structured knowledge distillation works.

Structured knowledge distillation can be applied to other structured output prediction tasks with FCN frameworks. In Section \ref{sec:Detec} and Section \ref{sec:depth}, we apply our distillation method to strong baselines in object detection and depth estimation tasks with 
minimum 
modifications.

\subsection{Semantic Segmentation}
\subsubsection{Implementation Details}
\noindent\textbf{Network structures.}
We
use the 
segmentation architecture of PSPNet \cite{zhao2017pyramid} with a ResNet$101$ \cite{He2016DeepRL} as the %
teacher
network $\mathsf{T}$.

We study recent 
compact networks, and employ several different architectures to verify the effectiveness of the distillation framework. 
We first
use 
ResNet$18$ as a basic student network and conduct ablation studies. 
Then, we employ the 
MobileNetV$2$Plus \cite{LightNet}, which is based on a pretrained MobileNetV$2$~\cite{Sandler2018MobileNetV2IR} model on the ImageNet dataset.
We also test ESPNet-C \cite{mehta2018espnet} and ESPNet \cite{mehta2018espnet} models,
 which are 
 also 
 lightweight models.

\noindent\textbf{Training setup.}
Unless otherwise specified, 
segmentation networks here using stochastic gradient descent (SGD) with the momentum ($0.9$) 
and the weight decay ($0.0005$) for $40000$ iterations. 
The learning rate is initialized %
to be 
$0.01$ and is multiplied by $(1-\frac{ iter}{ max-iter})^{0.9}$. 
We random 
crop 
the images into the resolution of $512 \times 512$ pixels as the training input. 
Standard 
data augmentation is 
applied during training, such as random scaling (from $0.5$ to $2.1$) and random flipping. 
We follow the settings in \cite{mehta2018espnet} to reproduce the results of ESPNet and ESPNet-C, and train the compact networks
with our distillation framework.

\subsubsection{Datasets}
\noindent\textbf{Cityscapes}. 
The Cityscapes dataset~\cite{Cordts2016Cityscapes}
is collected for urban scene understanding
and contains $30$ classes with only $19$ classes
used for evaluation. 
The dataset contains $5,000$ high quality 
finely annotated images and $20,000$ coarsely annotated images. The finely annotated images are divided into $2,975$, $500$,  $1,525$ images for training, validation and testing. We only use the finely annotated dataset in our experiments.

\noindent\textbf{CamVid}. 
The CamVid dataset~\cite{brostow2008segmentation} %
contains $367$ training and $233$ testing images. 
We evaluate the performance over $11$ different classes such as building, tree, sky, car, road, etc.

\noindent\textbf{ADE$20$K}. 
The ADE$20$K dataset~\cite{zhou2017scene} %
contains $150$ classes of  diverse scenes. 
The dataset is divided into $20K$/$2K$/$3K$ images for training, validation and testing.

\subsubsection{Evaluation Metrics}

We use the following metrics to evaluate the segmentation accuracy, model size and the efficiency.

The  
\emph{Intersection over Union (IoU)} score is calculated
as the ratio of interval and union between
the ground-truth mask and the predicted segmentation mask
for each class. 
We use the mean IoU of all classes ({mIoU}) to study the effectiveness of distillation. 
We also report the class IoU to study the effect of distillation on different classes.
\emph{Pixel accuracy} is the ratio of the pixels with the correct semantic labels to the overall pixels.

The \emph{model size} is represented by the number of network parameters.
and the \emph{complexity} is evaluated by the sum of floating point operations (FLOPs) in one forward on a fixed input size.

\subsubsection{Ablation Study}
\label{sec:ablation}
\noindent\textbf{The effectiveness of distillations}.
We 
examine 
the role of each component 
of our distillation system. 
The experiments are conducted on ResNet$18$ with its variant ResNet$18$ ($0.5$) representing a width-halved version of ResNet$18$ on the Cityscapes dataset. 
Table \ref{tab:ablition} reports
the results of different settings for the student net, which are 
the average results of the final epoch from three 
runs.

\renewcommand{\arraystretch}{1.2}
\begin{table}[b]
\centering
\footnotesize
\caption{The effect of different components of the loss in the proposed method. PI: pixel-wise distillation;
PA: pair-wise distillation;
HO: holistic distillation;
ImN:  initialized from the pretrain weight on the ImageNet.}

\begin{tabular}{l|c|c}
\hline
Method & Validation mIoU (\%)& Training mIoU (\%)  \\
\hline
Teacher&$78.56$&$86.09$\\
\hline
ResNet$18$ ($0.5$) & $55.37\pm0.25$ & $60.67\pm0.37$\\
+ PI &$57.07\pm0.69$&$62.33\pm0.66$\\
+ PI + PA & $61.52\pm0.09$&$66.03\pm0.07$\\
+ PI + PA + HO &$\bf{62.35}\pm0.12$&$\bf{66.72\pm0.04}$\\
\hline
ResNet$18$ ($1.0$) & $57.50\pm0.49$ & $62.98\pm0.45$\\
+ PI &$58.63\pm0.31$&$64.32\pm0.32$\\
+ PI + PA & $62.97\pm0.06$&$68.97\pm0.03$\\
+ PI + PA + HO &$\bf{64.68}\pm0.11$&$\bf{70.04}\pm0.06$\\
ResNet$18$ ($1.0$)&$69.10\pm0.21$&$74.12\pm0.19$\\
+ PI +ImN&$70.51\pm0.37$&$75.10\pm0.37$\\
+ PI + PA +ImN  &$71.78\pm0.03$&$77.66\pm0.02$\\
+ PI + PA + HO +ImN  &$\bf{74.08}\pm0.13$&$\bf{78.75\pm0.02}$\\
\hline
\end{tabular}
\label{tab:ablition}

\end{table}

From Table \ref{tab:ablition}, we can see that distillation 
can improve the performance of the student network, 
and structure distillation helps the student learn better. 
With the three distillation terms,
the improvements for ResNet$18$ ($0.5$), ResNet$18$ ($1.0$) and ResNet$18$ ($1.0$) initialized with pretrained weights on the ImageNet dataset
are $6.26\%$, $5.74\%$ and $2.9\%$, respectively,
which indicates that the effect of distillation is 
more pronounced for the smaller student network and networks without initialization with the weight pretrained from the ImageNet.  
Such an initialization is also 
able
to transfer the knowledge from another source (ImageNet). 
The best mIoU of the holistic distillation for ResNet$18$ ($0.5$) reaches $62.7\%$ on the validation set.

On the other hand, each distillation scheme leads to higher mIoU scores.
This implies that the three distillation schemes make
complementary contributions for better training the compact network.

\noindent\textbf{The affinity graph in pair-wise distillation}.
In this section, we discuss the impact of the connection range $\alpha$ and the granularity of each node $\beta$ in building the affinity graph. To calculate the pair-wise similarity among each pixel in the feature map will form the fully connected 
affinity graph, 
with the price of high computational complexity.
We fix the node to be one pixel, and 
vary 
the connection range $\alpha$ from the fully connected graph to local sub-graph. Then, we keep the connection range $\alpha$ to be fully 
connected, 
and use a local patch to denote each node to change the granularity $\beta$ from fine to coarse. The result are shown in Table \ref{tab:graph}. The results of different settings for the pair-wise distillation are the average results from three runs. We employ a ResNet$18$ ($1.0$) with the weight pretrained from ImageNet as the student network. All the experiments are performed with both pixel-wise distillation and pair-wise distillation, but the sizes of the affinity graph in pair-wise distillation %
vary.

From Table \ref{tab:graph}, we can see 
that 
increasing the connection range can help improve the distillation performance. With the fully connected graph, the student can achieve around $71.37\%$ mIoU. 
It appears that 
the best $\beta$ is $2 \times 2$, which is 
slightly  
better than the finest affinity graph, but the connections are significantly decreased. Using a small local patch to denote a node and calculate the affinity graph may form a more stable correlation between different locations. One can choose to use the local patch to 
decrease 
the number of the nodes, instead of decreasing the connection range for a better trade-off between efficiency and accuracy.

To include more structure information, we fuse pair-wise distillation items with different affinity graphs. Three pair-wise fusion
strategies 
are introduced: $\alpha$ fusion, $\beta$ fusion and 
feature level fusion. The details are shown in Table~\ref{tab:graph}. We can see that combining more affinity graphs may 
improve the performance, 
but also introduces extra computational cost during training. Therefore, we apply only one pair-wise distillation item in our methods.

\begin{center}
\begin{table}[htb]
\centering

\caption{The impact of the connection range and node granularity. The shape of the output feature map is $H'\times W'$. We can see that to keep a fully connected graph is more helpful in pair-wise distillation.}
\begin{threeparttable}
\begin{tabular}{c|c|c}
\hline
Method   & Validation mIoU(\%) &Connections \\ \hline
Teacher   & $78.56$               & $-$                                                            \\ \hline
Resnet18 (1.0)   & $69.10\pm0.21$           & $-$                                                            \\ \hline
\multicolumn{3}{c}{$\beta=1\times1$, $\alpha=$} \\
\hline
$W'/16\times H'/16$& $70.83\pm0.12$ & $(W'\times H')^2/2^8$   \\ 
$W'/8\times H'/8$         &$ 70.94\pm0.11$           & $(W'\times H')^2/2^6$                                                      \\ 
$W'/4\times H'/4$         & $71.09\pm0.07$           & $(W'\times H')^2/2^4$                                                      \\ 
$W'/2\times H'/2$         & $71.15\pm0.01$           & $(W'\times H')^2/4$                                                      \\ 
  $W\times H$ & {$71.37\pm0.12$}           & $(W'\times H')^2$                                                    \\ \hline
 \multicolumn{3}{c}{$\alpha=W'\times H'/\beta$, $\beta=$}   
\\\hline
 $2\times 2$         &$ \bf{71.78\pm0.03}$           & $(W'\times H')^2/2^4$                                                      \\  
$4\times 4$         & $71.24\pm0.18$           & $(W'\times H')^2/2^8$                                                      \\  
$8\times 8$         & $71.10\pm0.36$           & $(W'\times H')^2/2^{12}$                                                      \\  
 $16\times 16$ & {$71.11\pm0.14$}           & $(W'\times H')^2/2^{16}$ \\
$32\times 32$          & $70.94\pm0.23$           & $(W'\times H')^2/2^{20}$                                      \\ \hline
 \multicolumn{3}{c}{{Multi-level pair-wise distillations}}   
\\\hline
{ $\alpha$ Fusion\tnote{1}}        &$ 72.03\pm0.26$           &  $21* (W'\times H')^2/2^8$  \\
{ $\beta$ Fusion\tnote{2}}        &$ 71.91\pm0.17$           & $273*(W\times H)^2/2^{12}$\\
{ Feature-level Fusion\tnote{3}}        &$ \bf{72.18\pm0.12}$           & $3* (W'\times H')^2/2^4$                                           \\\hline
\end{tabular}

\begin{tablenotes}
\footnotesize
 \item[1] Different connection ranges fusion, with output feature size $H'\times W'$, $\beta=2\times2$ and $\alpha$ as $ W'/2\times H'/2$, $W'/4\times H'/4$ and $W'/8\times H'/8$, respectively. 
 \item[2] Different node granulates fusion, with output feature size $H'\times W'$, $\beta$ as $2\times2$, $4\times4$ and $8\times8$, respectively,
 and $\alpha$ as $ W'\times H'/\beta$. 
\item[3] Different feature levels fusion, with feature size $H'\times W'$, $2H'\times 2W'$, $4H'\times 4W'$,  $\beta$ as $2\times2$, $4\times4$ and $8\times8$, respectively, and $\alpha$ as maximum.
\end{tablenotes}
\end{threeparttable}
\label{tab:graph}
\end{table}
\end{center}

\begin{figure}[t!]
\begin{center}
\includegraphics[width=0.5\textwidth]{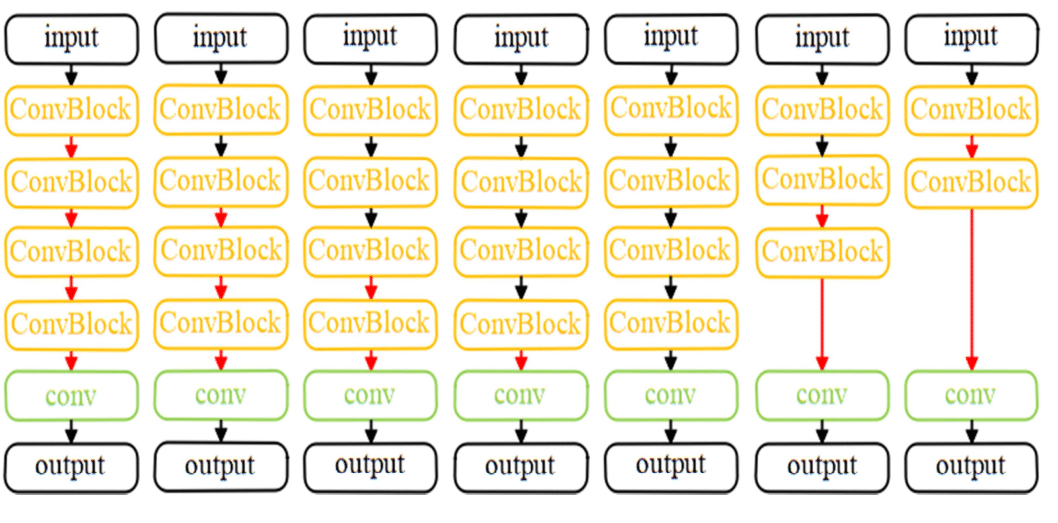}
\end{center}
\caption{We show $7$ different architectures of the discriminator. %
The red arrow represents %
a
self-attention layer. The orange block denotes a convolution block with stride $2$. We 
insert 
an average pooling layer to the output block to %
obtain 
the final score.}
\label{fig:dis_structure}
\end{figure}

\noindent\textbf{Adversarial training in holistic distillation}.
In this section, we illustrate that GAN is able to encode the holistic knowledge. 
Details of the discriminator are 
described in Section \ref{sec:ho}. 
The capability of discriminator 
would 
affect
the adversarial training, and we conduct experiments to discuss the impact of the discriminator's architecture. The results are shown in Table~\ref{Tab:d_structure}. We use $AnLm$ to represent the architecture of the discriminator with $n$ self-attention layers and $m$ convolution 
blocks 
with BN layers. The detailed structures can be seen in Figure~\ref{fig:dis_structure}, and the red arrows represent %
self-attention layers.

From %
Table~\ref{Tab:d_structure}, we can see adding self-attention layers can improve mIoU, and adding more self-attention layers %
does not change 
the results much.
We add $2$ self-attention blocks considering the performance, stability, and computational cost in our discriminator.  With the same self-attention layer, a deeper discriminator can help the adversarial training.

\begin{table}[htp]
\centering
\caption{We choose ResNet$18$ ($1.0$) as the example for the student net. An $AnLm$ index represents for $n$ self-attention layers with $m$ convolution blocks in the discriminator. The ability of discriminator affects the adversarial training.}
\begin{tabular}{c|c}
\hline
Architecture Index & Validation mIoU (\%) \\ \hline
 \multicolumn{2}{c}{Changing  self-attention layers}\\ \hline   
A4L4               & $73.46\pm0.02$           \\ \hline
A3L4               & $73.70\pm0.02$           \\ \hline
A2L4 (ours)         & $\bf{74.08\pm0.13} $           \\ \hline
A1L4              & $74.05\pm0.55$           \\ \hline
A0L4               & $72.85\pm0.01$           \\ \hline
\multicolumn{2}{c}{Removing  convolution blocks}\\ \hline
A2L4 (ours)         & $\bf{74.08\pm0.13} $           \\ \hline
A2L3               & $73.35\pm0.10$           \\ \hline
A2L2               & $72.24\pm0.42$           \\ \hline
\end{tabular}
\label{Tab:d_structure}
\end{table}

\begin{table*}[htp]
\centering
\caption{We choose ResNet$18$ ($1.0$) as the example for the student net. The numbers of class IoU with three different discriminator architectures are reported. Self-attention layers can significantly improve the accuracy of structured objects, such as truck, bus, train and motorcycle.}
\begin{tabular}{c|c|c|c|c|c|c|c|c|c|c}
\hline
Class            & mIoU           & road           & sidewalk       & building       & wall           & fence          & pole           & Traffic light  & Traffic sign   & vegetation     \\ \hline
D\_Shallow       &$72.28$           & $97.31$          & $80.07$          & $91.08$          & $36.86$          & $50.93$          & $62.13$          & $66.4$           & $76.77$          & $91.73$          \\ \hline
D\_no\_attention & $72.69$           & $\bf{97.36}$ & $\bf{80.22}$ & $\bf{91.62}$ &$\bf{45.16}$ & $\bf{56.97}$ &$ 62.23$          &$\bf{68.09}$ & $\bf{76.38}$ & $\bf{91.94}$ \\ \hline
D\_Ours       & $74.10 $          & $97.15$          & $79.17$          & $91.60$           & $44.84$          & $56.61$          & $\bf{62.37}$ & $67.37$          & $76.34$          & $91.91$          \\ \hline
class            & terrain        & sky            & person         & rider          & car            & truck          & bus            & train          & motorcycle     & bicycle        \\ \hline
D\_Shallow       & $60.14$          & $93.76$          & $79.89$          & $55.32$          & $93.45$          & $69.44$          & $73.83$          & $69.54$          & $48.98$          & $75.78$          \\ \hline
D\_no\_attention & $\bf{62.98}$ & $\bf{93.84}$ & $\bf{80.1}$  & $\bf{57.35}$ & $93.45$          & $68.71$          & $75.26$          & $56.28$          & $47.66$          & $75.59$          \\ \hline
D\_Ours      & $58.67$          & $93.79$          & $79.9$           & $56.61$          &$ \bf{94.3}$  & $\bf{75.83}$ & $\bf{82.87}$ & $\bf{72.03}$ & $\bf{50.89}$ & $\bf{75.72}$ \\ \hline
\end{tabular}
\label{fig:class_acc_d}
\end{table*}

To verify the effectiveness of adversarial training, we further explore the
capability of three typical discriminators: the shallowest one (D\_Shallow, i.e.,  A2L2), the one without attention layer (D\_no\_attention, i.e., A0L4) and ours (D\_Ours, i.e., A2L4).  The IoUs for different classes are listed in Table~\ref{fig:class_acc_d}. It is 
clear that self-attention layers can help the discriminator better capture the structure, thus the accuracy of the students with the structure objects is improved.

In the adversarial training, the student, a.k.a.\  the generator, 
tries 
to learn the distribution of the real samples (output of the teacher).
We apply the Wasserstein distance to transfer the 
distance 
between 
two distributions into a more intuitive score, and the score are highly 
relevant 
to the quality of the segmentation maps. We use a well-trained discriminator $\mathsf{D}$ (A2L4) to evaluate the score of a segmentation map. For each image, we feed five segmentation maps, output by the teacher net, the student net w/o holistic distillation, and the student nets w/ holistic distillation under three different discriminator architectures (listed in Table \ref{fig:class_acc_d}) into the discriminator $\mathsf{D}$,
and compare the distribution of embedding scores.
We evaluate on the validation set and calculate the average score difference between different student nets and the teacher net, the results are shown in Table \ref{tab:score_diff}.  With holistic distillation, the segmentation maps produced from student net can achieve a similar score to the teacher, 
indicating that GAN helps distill the holistic structure knowledge. 

\begin{table}[htp]
\centering
\caption{We choose ResNet$18$ ($1.0$) as the example for the student net. The embedding score difference and mIoU on the validation set of Cityscapes are presented.
}
\begin{tabular}{c|c|c}
\hline
Method              & Score difference & mIoU  \\ \hline
Teacher             & $0$                & $78.56$ \\ \hline
Student w/o D       & $2.28$             & $69.21$ \\ \hline
w/ D\_no\_attention & $0.23 $            & $72.69$ \\ \hline
w/ D\_shallow       & $0.46  $           & $72.28$ \\ \hline
w/ D\_ours          & $0.14$             & $74.10$ \\ \hline
\end{tabular}
\label{tab:score_diff}
\end{table}

\begin{figure}[t]
\centering 
\includegraphics[width=0.415\textwidth]{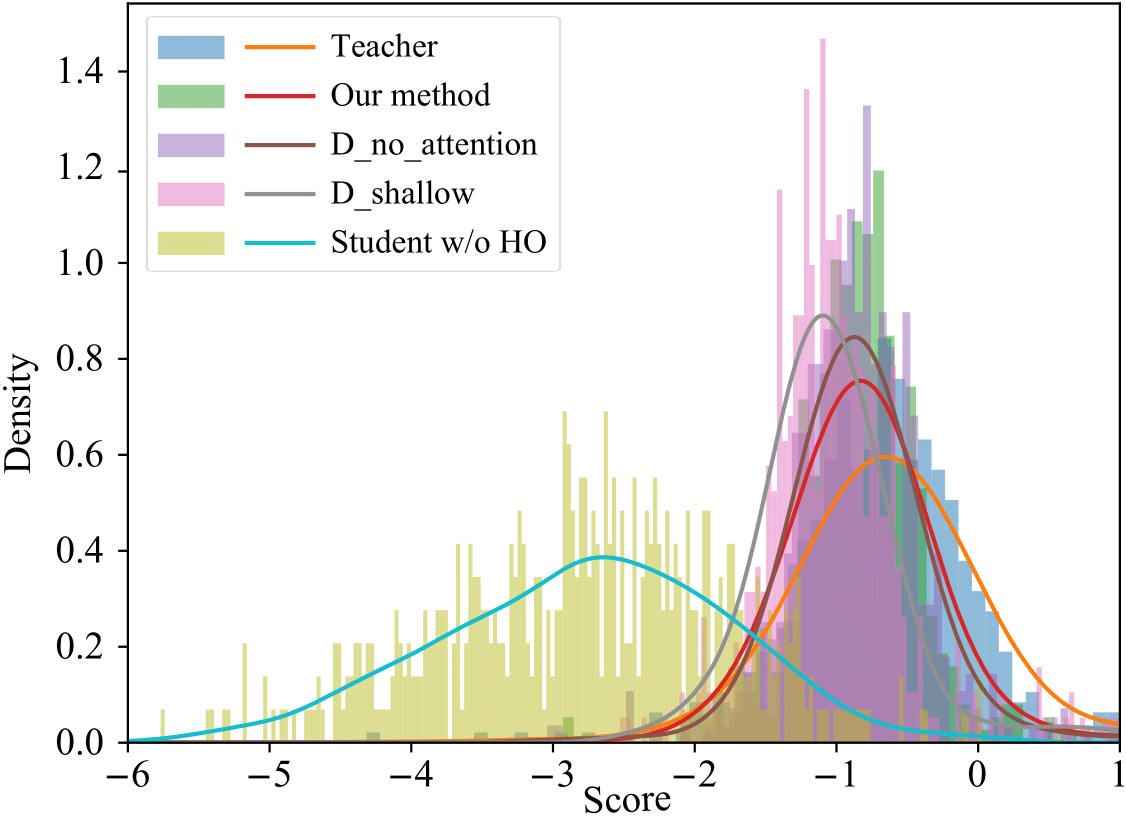}
\caption{The score distribution of segmentation maps generated by different student nets evaluated by a well-trained discriminator. With adversarial training, score distributions of segmentation maps become %
closer 
to
the teacher (the orange one); and our method (the red one) is the %
closest 
one 
to
the teacher.}
\label{fig:score}
\end{figure}

We also draw a histogram to show score distributions of segmentation maps across the validation set in Figure \ref{fig:score}.
The well-trained $\mathsf{D}$ can assign a higher score to high quality segmentation maps, and the three student nets with the holistic distillation can generate segmentation maps with higher scores and better quality. Adding self-attention layers and more convolution blocks help the student net to imitate the distribution of the teacher net, and attain 
better performance.

\begin{figure*}[t]
\setlength{\belowcaptionskip}{-0.2cm}
\centering  
\subfloat[Bus]{
\includegraphics[width=0.49\textwidth]{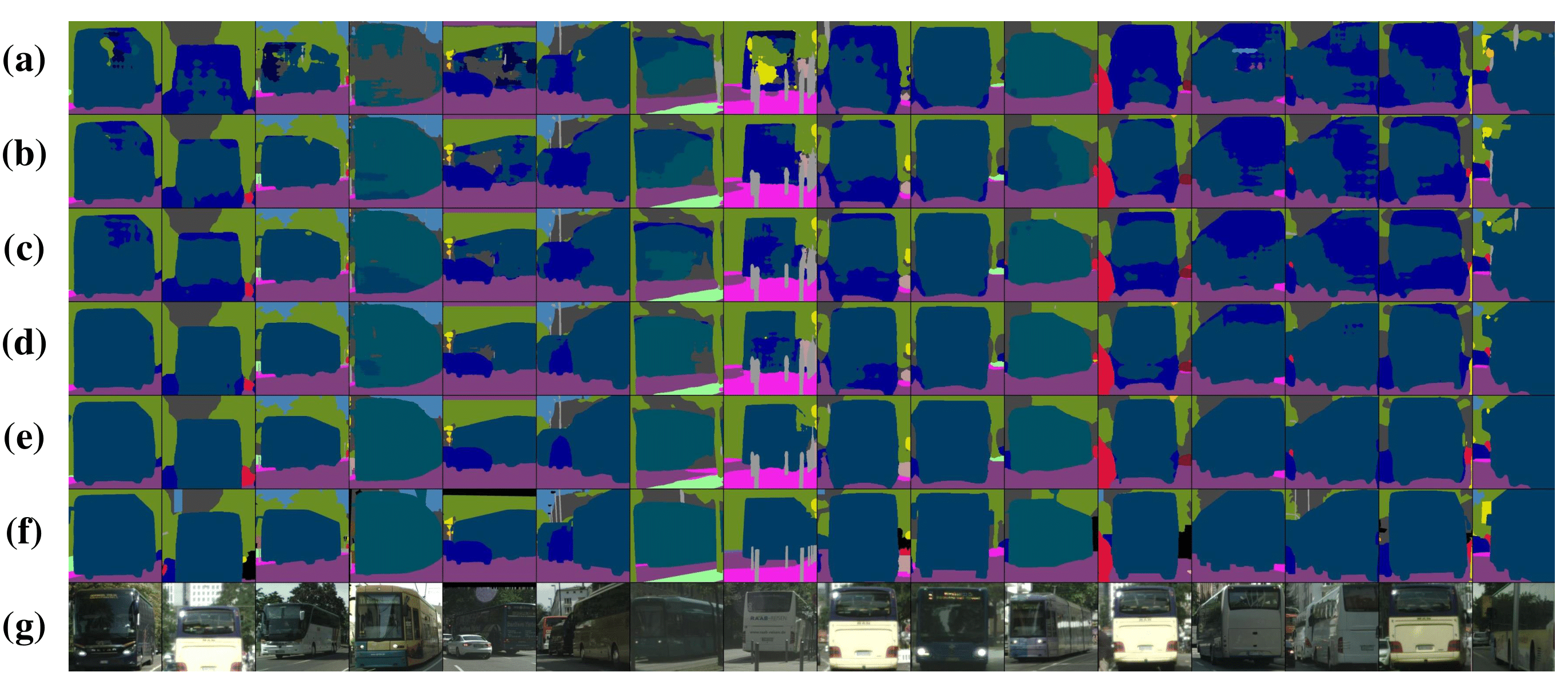}}
\subfloat[Truck]{
\includegraphics[width=0.49\textwidth]{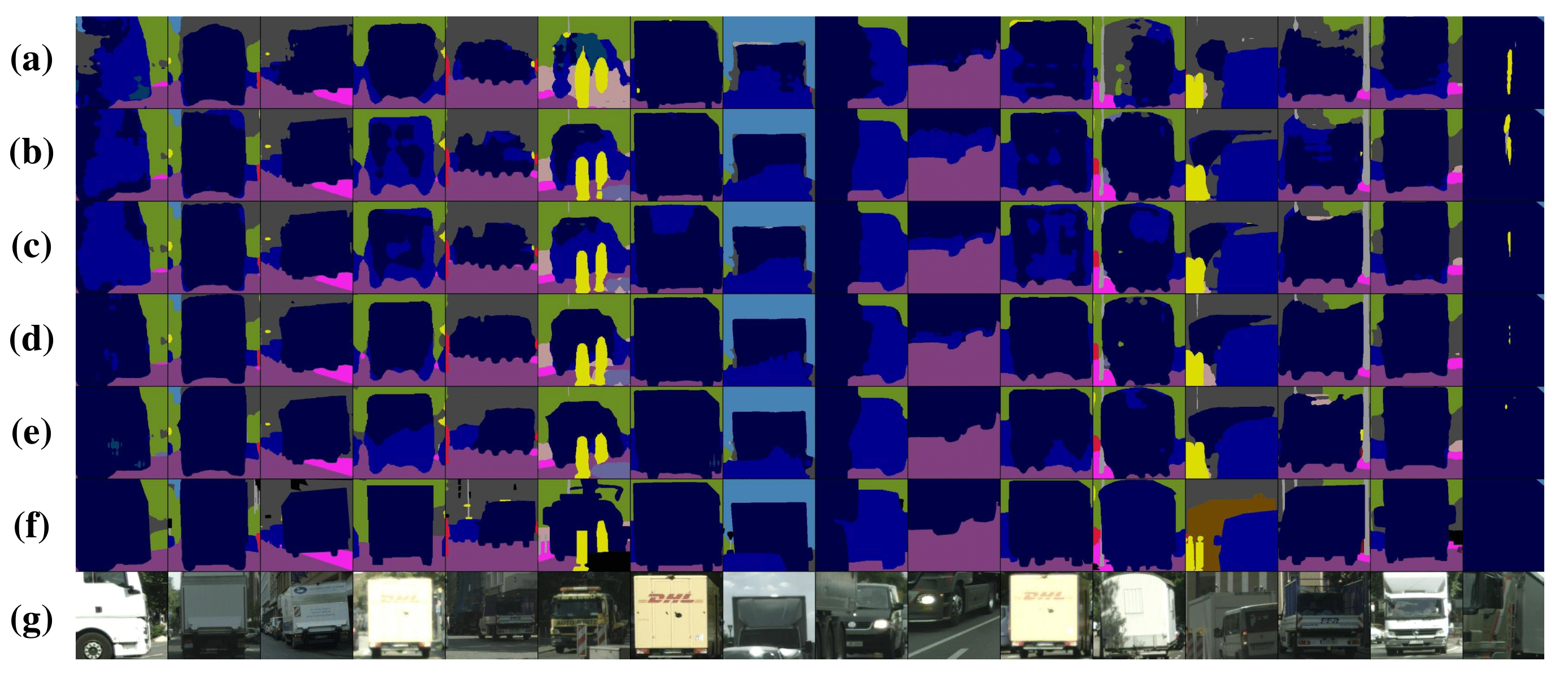}}
\caption{Segmentation results for structured objects with ResNet$18$ ($1.0$) trained with different discriminators. (a) W/o holistic distillation, (b)  W/ D{\_}shallow, (c) W/ D{\_}no{\_}attention, (d) Our method, (e) Teacher net, (f) Ground truth, (g) Image. One can see a strong discriminator can help the student learn structure objects better. With the attention layers, labels of the objects are more consistent.}

\label{fig:structure objects}
\end{figure*}

\noindent\textbf{Feature and local pair-wise distillation}.
We compare 
a few 
variants of the pair-wise distillation:
\begin{itemize}
    \item Feature distillation by MIMIC  \cite{romero2014fitnets, Li2017MimickingVE}: We follow \cite{Li2017MimickingVE} to align the features 
    of each pixel between $\mathsf{T}$ and $\mathsf{S}$
    through a $1 \times 1$ convolution layer to match the dimension of the feature
    \item Feature distillation by attention transfer \cite{Zagoruyko2016PayingMA}: 
    We aggregate the response maps into an 
    attention map (single channel),
    and then transfer the attention map from the teacher to the student.
    \item Local pair-wise distillation \cite{xieimproving}: This method can be seen as a special case of our pair-wise distillation, which only cover a small sub-graph ($8$-neighborhood pixels for each node).
\end{itemize}

\begin{table}[t]
\footnotesize
\caption{
Comparison of
feature transfer MIMIC~\cite{romero2014fitnets, Li2017MimickingVE},
attention transfer~\cite{Zagoruyko2016PayingMA},
and local pair-wise distillation~\cite{xieimproving} 
against 
our
pair-wise distillation.
The segmentation is evaluated by mIoU (\%).
PI: pixel-wise distillation. 
MIMIC: using a $1\times 1$ convolution layer for feature distillation. 
AT: attention transfer for feature distillation.
LOCAL: local similarity distillation method.
PA: our pair-wise distillation. 
ImN: initializing the network from the weights pretrained on ImageNet dataset.}

\begin{center}
\begin{tabular}{l|d{2.2}d{2.2}}
\hline
Method&\multicolumn{1}{c}{ResNet$18$ ($0.5$)} & \multicolumn{1}{c}{ResNet$18$ ($1.0$) + ImN}\\
\hline
w/o distillation   &$55.37$        & $69.10$  \\
+ PI & $57.07$&$70.51$\\
+ PI + MIMIC &$ 58.44$&$71.03$\\
+ PI + AT & $57.93$&$70.70$\\
+ PI + LOCAL& $58.62$&$70.86$\\
+ PI + PA& $\bf{61}$.$\bf{52}$&$\bf{71}$.$\bf{78}$\\
\hline
\end{tabular}
\end{center}

\label{tab:compare}
\end{table}
We replace our pair-wise distillation 
by the above three distillation schemes to verify the effectiveness 
of our pair-wise distillation.
From Table \ref{tab:compare}, we can see that our pair-wise distillation method outperforms all the other distillation methods. 
The superiority over feature distillation schemes
(MIMIC \cite{Li2017MimickingVE} and attention transfer \cite{Zagoruyko2016PayingMA}, 
which transfers the knowledge for each pixel separately)
may be due the fact 
that we transfer the structured knowledge
other than aligning the feature for each individual pixel. 
The superiority 
over 
the local pair-wise distillation shows the effectiveness
of our fully connected pare-wise distillation, which is able to 
transfer the 
overall 
structure information other than a local boundary information~\cite{xieimproving}.

\begin{table}[t]
\footnotesize
\caption{The segmentation results on
the testing, 
validation (Val.)
set of Cityscapes. 
}
\begin{center}
\setlength{\tabcolsep}{2mm}{
\begin{threeparttable}

\begin{tabular}{l|cccc}
\hline
Method&\#Params (M)&FLOPs (B) &Test \tnote{\S} & Val.\\
\hline
\multicolumn{5}{c}{Current state-of-the-art results}\\
\hline
ENet \cite{paszke2016enet} \tnote{\dag}&$0.3580$&$3.612$&$58.3$&n/a\\
ERFNet \cite{yu2015multi} \tnote{\ddag}&$2.067$&$25.60$&$68.0$&n/a\\
FCN \cite{shelhamer2017fully} \tnote{\ddag}&$134.5$&$333.9$&$65.3$&n/a\\
RefineNet \cite{lin2019refinenet} \tnote{\ddag}&$118.1$&$525.7$&$73.6$&n/a\\
OCNet  \cite{Yuan2018OCNetOC}\tnote{\ddag} &$62.58$&$548.5$&$80.1$ &n/a\\
PSPNet \cite{zhao2017pyramid} \tnote{\ddag}  &$70.43$&$574.9$&$78.4$&n/a\\
\hline
\multicolumn{5}{c}{Results w/ and w/o distillation schemes}\\
\hline
MD \cite{xieimproving} \tnote{\ddag}&$14.3$5&$64.48$&n/a&$67.3$\\
MD (Enhanced) \cite{xieimproving} \tnote{\ddag}&$14.35$&$64.48$&n/a&$71.9$\\
\hline
ESPNet-C \cite{mehta2018espnet} \tnote{\dag}&$0.3492$&$3.468$&$51.1$&$53.3$\\
ESPNet-C (ours) \tnote{\dag}&$0.3492$&$3.468$ &$57.6$ &$59.9$  \\
\hline
ESPNet \cite{mehta2018espnet} \tnote{\dag}&$0.3635$&$4.422$&$60.3$&$61.4$\\
ESPNet (ours) \tnote{\dag}&$0.3635$&$4.422$&$62.0$&$63.8$ \\
\hline
ResNet$18$ ($0.5$) \tnote{\dag}&$3.835$&$33.35$&$54.1$&$55.4$\\
ResNet$18$ ($0.5$) (ours) \tnote{\dag}&$3.835$&$33.35$&$61.4$&$62.7$  \\
\hline
ResNet18 (1.0) \tnote{\ddag}& $15.24$&$128.2$&$67.6$&$69.1$\\
ResNet18 (1.0) (ours) \tnote{\ddag}&$15.24$&$128.2$&$73.1$&$75.3$  \\
\hline
MobileNetV$2$Plus \cite{LightNet} \tnote{\ddag}&$8.301$&$86.14$&$68.9$&$70.1$\\
MobileNetV$2$Plus (ours) \tnote{\ddag}&$8.301$&$86.14$&$74.0$&$74.5$ \\

\hline
\end{tabular}

\begin{tablenotes}
\footnotesize
\item[\dag] Train from scratch
\item[\ddag] Initialized from the weights pretrained on ImageNet
\item[\S] We select a best model along training on validation set to submit to the leader board. All our models are test on single scale. Some teacher networks are test on multiple scales, such as OCNet and PSPNet. 
\end{tablenotes}
\end{threeparttable}}
\label{tab:all}
\end{center}
\end{table}

\begin{figure*}[htb]
\begin{center}
\includegraphics[width=1\textwidth]{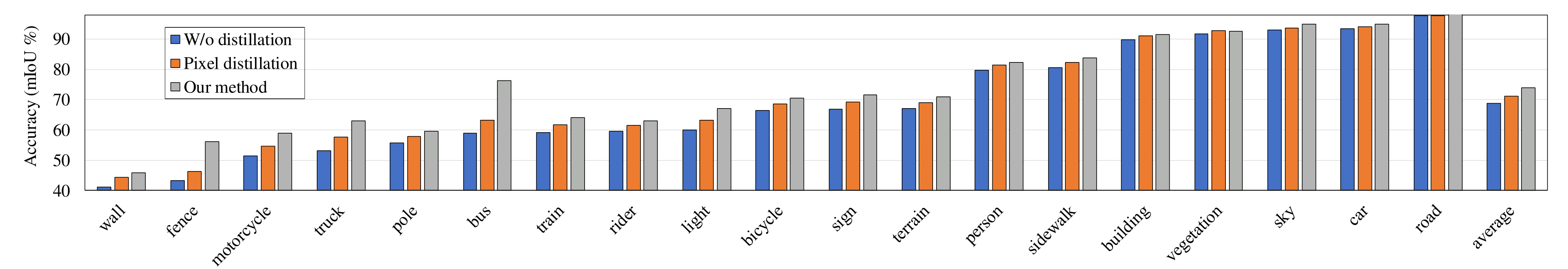}
\end{center}

\caption{ Illustrations of the effectiveness
of structured distillation schemes 
in terms of class IoU scores 
using 
the network MobileNetV2Plus \cite{LightNet} 
on 
the Cityscapes test set. 
Both pixel-level and structured distillation are helpful for improving the performance
especially for the hard classes with low IoU scores.
The improvement from structured distillation is more significant
for structured objects, such as bus and truck. }
\label{fig:class}
\end{figure*}
\begin{figure*}[htb]
\centering  %
\subfloat[Image]{
\label{Fig.sub.1}
\includegraphics[width=0.2\textwidth]{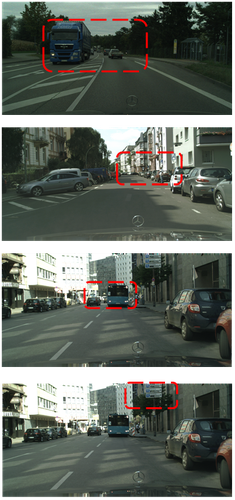}}
\subfloat[W/o distillation]{
\label{Fig.sub.2}
\includegraphics[width=0.2\textwidth]{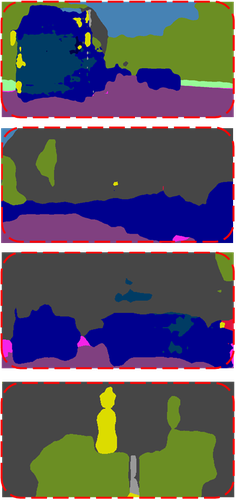}}
\subfloat[Pixel-wise distillation]{
\label{Fig.sub.3}
\includegraphics[width=0.2\textwidth]{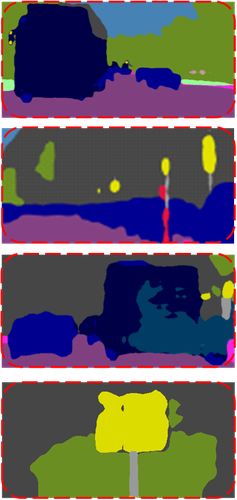}}
\subfloat[Our method]{
\label{Fig.sub.4}
\includegraphics[width=0.2\textwidth]{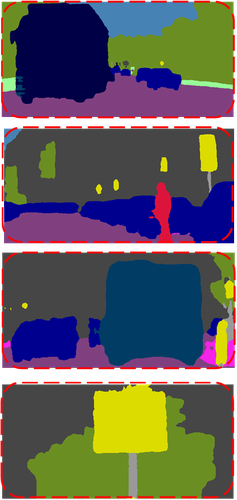}}
\subfloat[Ground truth]{
\label{Fig.sub.5}
\includegraphics[width=0.2\textwidth]{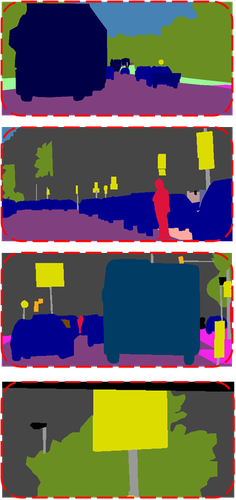}}

\caption{Qualitative results on the Cityscapes testing set 
produced from MobileNetV2Plus: 
(a) initial images,
(b) w/o distillation,
(c) only w/ pixel-wise distillation,
(d) Our distillation schemes: both pixel-wise and structured distillation schemes.
The segmentation map in the red box
about four structured objects: trunk, person, bus and traffic sign
are zoomed in. 
One can see that 
the structured distillation method (ours) produces more consistent labels.}
\label{Fig.main}
\end{figure*}

\subsubsection{Segmentation Results}
\label{sec:sota}
\noindent\textbf{Cityscapes}. 
We apply our structure distillation method to several compact networks: 
MobileNetV2Plus \cite{LightNet} which is based on a MobileNetV2 model, 
ESPNet-C \cite{mehta2018espnet} and ESPNet \cite{mehta2018espnet}
which are carefully designed for mobile applications. 
Table \ref{tab:all}
presents the segmentation accuracy, the model complexity and the model size.
FLOPs\footnote{The FLOPs is calculated with the pytorch version implementation \cite{FLOPs}.
}
is calculated on the resolution of $512\times1024$ pixels to evaluate the complexity. 
\#Params is the number of network parameters.
We can see that
our distillation approach can 
improve the results
over $5$ compact networks:
ESPNet-C and ESPNet~\cite{mehta2018espnet},
ResNet$18$ ($0.5$),
ResNet$18$ ($1.0$),
and MobileNetV2Plus~\cite{LightNet}.
For the networks without pre-training, such as 
ResNet$18$ ($0.5$)  and ESPNet-C, 
the improvements 
are very significant with
$7.3\%$ and $6.6\%$, respectively. 
Compared with MD (Enhanced) \cite{xieimproving}
that uses the pixel-wise and local pair-wise distillation schemes
over MobileNet,
our approach with the similar network MobileNetV2Plus 
achieves higher segmentation quality 
($74.5$ vs.\  $71.9$
on the validation set)
with a little higher computation complexity
and much smaller model size.

Figure \ref{fig:class} shows the IoU scores for each class
over MobileNetV2Plus.
Both the pixel-wise and structured distillation schemes
improve the performance,
especially for the categories 
with low IoU scores.
In particular,
the structured distillation (pair-wise and holistic) has significant improvement
for structured objects, e.g.,  $17.23\%$ improvement for Bus 
and $10.03\%$ for Truck. 
The qualitative segmentation results in Figure \ref{Fig.main} 
visually demonstrate the effectiveness of our structured distillation
for structured objects, such as trucks, buses, persons, and traffic signs.

\begin{figure}[t]
\setlength{\belowcaptionskip}{-0.2cm}
\centering  
\subfloat[Image]{
\includegraphics[width=0.11\textwidth]{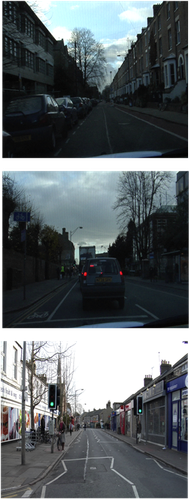}}
\subfloat[W/o dis.]{
\includegraphics[width=0.11\textwidth]{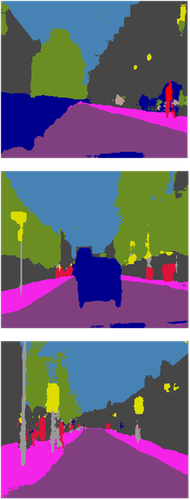}}
\subfloat[Our method]{
\includegraphics[width=0.11\textwidth]{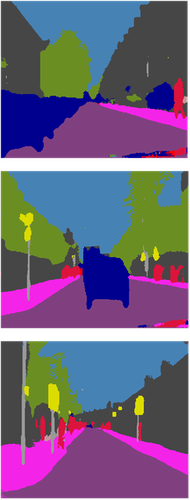}}
\subfloat[Ground truth]{
\includegraphics[width=0.11\textwidth]{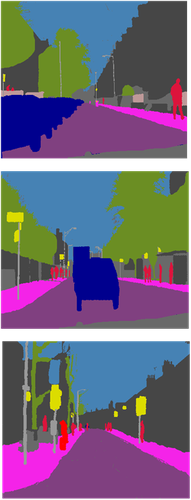}}

\caption{Qualitative results on the CamVid test set produced from ESPNet. W/o dis. represents for the baseline student network trained without distillation.}

\label{fig:camvid}
\end{figure}

\noindent\textbf{CamVid}.
Table \ref{tab:all_camvid}
shows the performance
of the student networks
w/o and w/ our distillation schemes
and state-of-the-art results. 
We train and evaluate the student networks
w/ and w/o distillation
at the resolution $480 \times 360$ following the setting of ENet. 
Again we can see that the distillation scheme improves 
the performance. 
Figure \ref{fig:camvid} shows some samples on the CamVid test set w/o and w/ the distillation produced from ESPNet.
\begin{table}[t]
\footnotesize
\caption{The segmentation performance on the test set of CamVid. `ImN': ImageNet dataset;
`unl': unlabeled street scene dataset sampled from Cityscapes. }

\begin{center}
\begin{tabular}{l|ccc}
\hline
Method&Extra data&\multicolumn{1}{c}{mIoU (\%)} & \#Params (M)\\
\hline
ENet\cite{paszke2016enet}&no&$51.3$&$0.3580$\\
FC-DenseNet$56$\cite{drozdzalone}&no&$58.9$&$1.550$\\
SegNet\cite{badrinarayanan2017segnet}&ImN&$55.6$&$29.46$\\
DeepLab-LFOV\cite{chen2014semantic}&ImN&$61.6$&$37.32$\\
FCN-$8$s\cite{shelhamer2017fully}&ImN&$57.0$&$134.5$\\

\hline

ESPNet-C\cite{mehta2018espnet}&no&$56.7$&~\\
ESPNet-C (ours)&no&$60.3$&$0.3492$\\
ESPNet-C (ours)&unl&$64.1$&~\\
\hline
ESPNet\cite{mehta2018espnet}&no&$57.8$&\\
ESPNet (ours)&no&$61.4$&$0.3635$\\
ESPNet (ours)&unl&$65.1$&\\
\hline
ResNet$18$ &ImN&$70.3$&\\
ResNet$18$ (ours)&ImN&$71.0$&$15.24$\\
ResNet$18$ (ours)&ImN+unl&$72.3$&\\
\hline
\end{tabular}
\end{center}

\label{tab:all_camvid}
\end{table}

\begin{figure}[t]
\begin{center}
\includegraphics[width=0.5\textwidth]{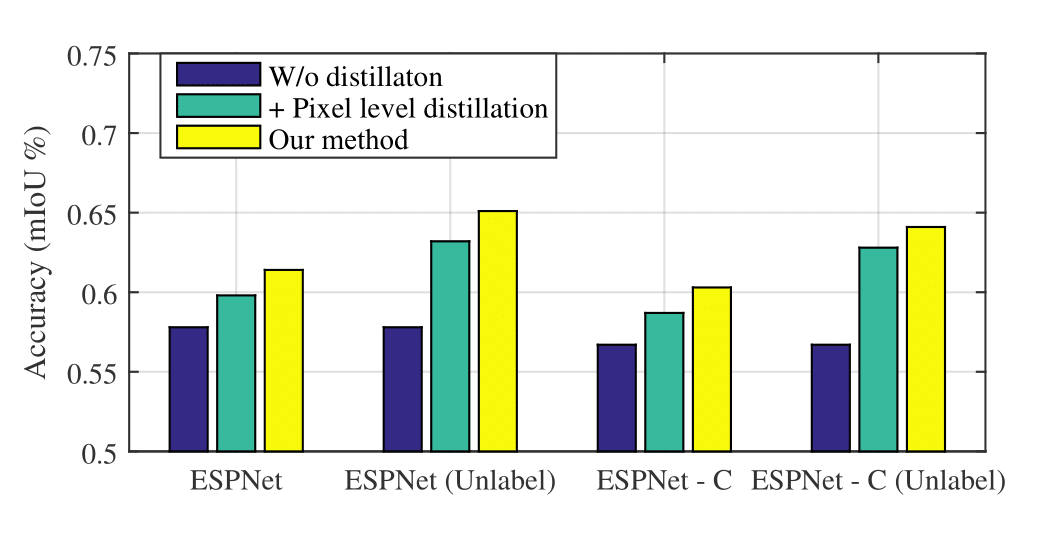}
\end{center}

\caption{The effect of structured distillation on CamVid. This figure shows that distillation can improve the results 
in two cases: trained over only the labeled data and 
over both the labeled and extra unlabeled data.}
\label{fig:unlabel}
\end{figure}

We also conduct an experiment 
by using an extra unlabeled dataset,
which contains $2000$ unlabeled street scene images
collected from the Cityscapes dataset,
to show that the distillation schemes
can transfer the knowledge of the unlabeled images.  
The experiments are done with ESPNet and ESPNet-C. 
The loss function is almost the same 
except that there is no cross-entropy loss over
the unlabeled dataset. 
The results are shown in Figure \ref{fig:unlabel}. 
We can see that our distillation method with the extra unlabeled data can significantly improve mIoU of ESPNet-C and ESPNet for $13.5\%$ and $12.6\%$. 

\noindent\textbf{ADE$20$K}.
The ADE$20$K dataset is a very challenging dataset 
and contains $150$ categories of objects. 
Note that the number of pixels belonging to difference categories in this dataset 
is very imbalanced. 

We report the results for ResNet$18$ and the MobileNetV$2$
which are trained with the initial weights pretrained 
on the ImageNet dataset, 
and ESPNet which is trained from scratch in Table \ref{tab:all_ADE}. We follow the same training scheme in \cite{xiao2018unified}. All results are tested on single scale. 
For ESPNet, with our distillation, 
we can see that the mIoU score is improved by $3.78\%$,
and it achieves
a higher accuracy with %
fewer 
parameters 
compared to SegNet. 
For ResNet$18$ and MobileNetV$2$, after the distillation, 
we 
achieve 
$2.73\%$ improvement 
over the one without distillation 
reported in \cite{xiao2018unified}.

\begin{table}[t]
\footnotesize

\caption{The mIoU and pixel accuracy on validation set of ADE$20$K.}

\begin{center}
\setlength{\tabcolsep}{1mm}
{
\begin{tabular}{l|ccc}
\hline
Method&mIoU(\%)&Pixel Acc. (\%)&\#Params (M)\\
\hline

SegNet \cite{badrinarayanan2017segnet}&$21.64$&$71.00$&$29.46$\\
DilatedNet$50$ \cite{xiao2018unified}&$34.28$&$76.35$&$62.74$\\
PSPNet (teacher) \cite{zhao2017pyramid}&$42.19$&$80.59$&$70.43$\\
FCN \cite{shelhamer2017fully}&$29.39$&$71.32$&$134.5$\\
\hline
ESPNet \cite{mehta2018espnet}&$20.13$&$70.54$&$0.3635$\\
ESPNet (ours)&$24.29$&$72.86$&$0.3635$\\

\hline
MobileNetV$2$ \cite{xiao2018unified} &$34.84$&$75.75$&$2.17$\\
MobileNetV$2$ (ours)&$38.58$&$79.78$&$2.17$\\
\hline
ResNet$18$ \cite{xiao2018unified}&$33.82$&$76.05$&$12.25$\\
ResNet$18$ (ours)&$36.60$&$77.97$&$12.25$\\
\hline
\end{tabular}
}
\end{center}
\label{tab:all_ADE}

\end{table}

\begin{figure}[t]
\setlength{\belowcaptionskip}{-0.2cm}
\centering  
\subfloat[Image]{
\includegraphics[width=0.13\textwidth]{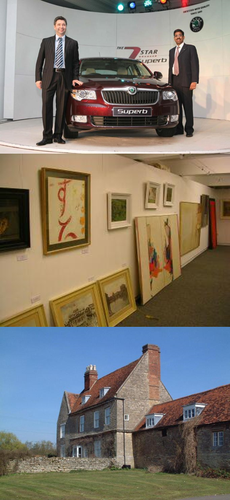}}
\subfloat[w/o dis.]{
\includegraphics[width=0.13\textwidth]{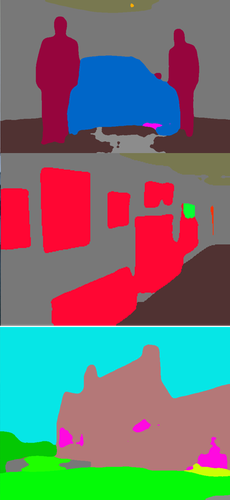}}
\subfloat[Our method]{
\includegraphics[width=0.13\textwidth]{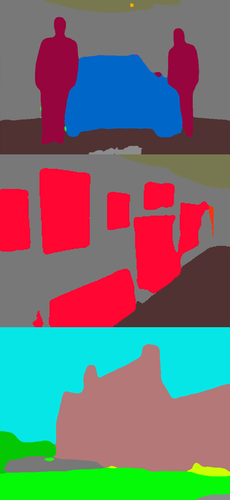}}

\caption{Qualitative results on the ADE$20$K produced from MobileNetV$2$. W/o dis.\  represents for the baseline student network trained without distillation.}

\label{fig:ADE20k}
\end{figure}

\subsection{Depth Estimation}
\label{sec:depth}

\subsubsection{Implementation Details}

\noindent\textbf{Network structures.}
We use the same model described in~\cite{wei2019enforcing} with the ResNext$101$ backbone as our teacher model, and replace the backbone with MobileNetV$2$ as the compact model.

\noindent\textbf{Training details.}
We train the student net using the crop size $385 \times 385$ by mini-batch stochastic gradient descent (SGD) with batchsize of $12$. The initialized learning rate is $0.001$  and is multiplied by $(1-\frac{iter}{max-iter})^{0.9}$. For both 
w/ and w/o distillation methods, the training epoch is $200$.

\subsubsection{Dataset}
\noindent\textbf{NYUD-V2.} The \textbf{NYUD-V2} dataset contains $1449$ annotated indoor images, in which $795$ images are 
for training and others are for testing. The image size is $640 \times 480$. 
 Some methods have  sampled more images from the video sequence of NYUD-V2 to form a \textbf{Large-NYUD-V2} to further improve the performance. 
Following~\cite{wei2019enforcing}, we 
conduct 
ablation studies on the small dataset and also apply the distillation method on current state-of-the-art real-time depth models trained with Large-NYUD-V2 to verify the effectiveness of the structured knowledge distillation.

\subsubsection{Evaluation Metrics}
We follow previous methods~\cite{wei2019enforcing} to evaluate the performance of monocular depth estimation quantitatively based on following metrics: mean absolute relative error (rel), mean $\log_{10}$ error ($\log_{10}$), root mean squared error (rms) , and the accuracy under threshold ($\delta_{i} < 1.25^{i}, i=1, 2, 3$). 
\subsubsection{Results}
\noindent\textbf{Ablation studies.}
We compare the pixel-wise distillation and the structured knowledge distillation in this section. In the dense classification problem, e.g., semantic segmentation, the output logits of the teacher is a soft distribution of all the classes, which contain the relations among different classes. Therefore, directly transfer the logits from  teacher  models from the compact ones 
at the 
pixel level can help improve the performance. Different from semantic segmentation, as the depth map are 
real 
values, the output of the teacher %
is often
not as accurate as ground truth labels. In the experiments, we find that adding pixel-level distillation hardly improves the accuracy in the depth estimation task.
Thus, 
we only use the structured knowledge distillation in depth estimation task.

To verify that the distillation method can further improve the accuracy with unlabeled data, we use $30K$ image sampled from the video sequence of NYUD-V2 without the depth map. The results are shown in Table \ref{tab:depth_abl}.
We can see that the structured knowledge distillation performs better than pixel-wise distillation, and adding extra  unlabelled data can  further improve the accuracy.
\begin{table}[t!]
\centering
\caption{\textbf{Depth estimation}. Relative error on 
the 
NYUD-V2 test dataset. `Unl'  means Unleblled data sampled from the large video sequence. The pixel-level distillation alone can not  improve the accuracy.
Therefore we only use structured-knowledge distillation in the depth estimation task.}
\begin{tabular}{c|c|c|c|c|c}
\hline
Method&Baseline&PI        & +PA        & +PA +HO        & +PA +HO +Unl      \\ \hline
 rel&$0.181$  & $0.183$      &   $0.175$        &   $0.173$        &    $\bf{0.160}$       \\ \hline

\end{tabular}
\vspace{0.1cm}

\label{tab:depth_abl}
\end{table}

\noindent\textbf{Comparison with  state-of-the-art.}
We apply the distillation method
to a few 
state-of-the-art %
lightweight 
models for depth estimation. Following~\cite{wei2019enforcing}, we train the student net on Large-NYUD-V2 with the same constraints in~\cite{wei2019enforcing} as our baseline, and achieve $13.5$ 
in the metric
`rel'. Following the same training setups, with the structured knowledge distillation terms, we further improve the strong baseline, and achieve a relative error (rel) of $13.0$. %
In Table~\ref{table:errors cmp on NYUD-V2}, we list the model parameters and %
accuracy of
a few
state-of-the-art large models along with some real-time models, indicating 
that 
the structured knowledge distillation works on a strong baseline.

\begin{table*}[htb]
\centering
\caption{Depth estimation results and model parameters on NYUD-V2 test dataset. With the structured knowledge distillation, the performance is improved over all evaluation metrics.}
\begin{tabular}{l|l c|ccccccc}
\toprule[1pt]
\multirow{2}{*}{Method} & backbone& \#Params (M)& \textbf{rel} & \textbf{log10} & \textbf{rms} & $\boldsymbol{\delta_{1}}$ & $\boldsymbol{\delta_{2}}$ & $\boldsymbol{\delta_{3}}$ \\
                      &&  & \multicolumn{3}{c}{Lower is better}         & \multicolumn{3}{c}{Higher is better} \\ \hline
Laina et al.~\cite{laina2016deeper}&ResNet$50$  &     $60.62$     & $0.127$        & $0.055$          & $0.573$        & $0.811$      & $0.953$      & $0.988$      \\
DORN~\cite{fu2018deep}             &ResNet$101$ & $105.17$     & $0.115$        & $0.051$          & $0.509$        & $0.828$      & $0.965$      & $0.992$      \\ 
AOB~\cite{hu2019revisiting}&SENET-$154$&$149.820.181$&$0.115$&$0.050$&$0.530$&$0.866$&$0.975$&$0.993$\\
VNL (teacher)~\cite{wei2019enforcing} &ResNext$101$ &$86.24$  & $\bf{0.108}$    & $\bf{0.048}$   & $\bf{0.416}$   & $\bf{0.875}$           & $\bf{0.976}$    &  $\bf{0.994}$          \\ \hline
\hline
CReaM~\cite{spek2018cream}&-&$\bf{1.5}$&$0.190$&-&$0.687$&$0.704$&$0.917$&$0.977$\\
RF-LW~\cite{nekrasov2018real}&MobileNetV$2$&$3.0$&$0.149$&-&$0.565$&$0.790$&$0.955$&$0.990$\\
VNL (student)&MobileNetV$2$&$2.7$&$0.135$&$0.060$&$0.576$&$0.813$&$0.958$&$0.991$\\
VNL (student) w/ distillation&MobileNetV$2$&$2.7$&$\bf{0.130}$&$\bf{0.055}$&$\bf{0.544}$&$\bf{0.838}$&$\bf{0.971}$&$\bf{0.994}$\\
\hline
\end{tabular}\newline

\label{table:errors cmp on NYUD-V2}
\end{table*}

\subsection{Object Detection}
\label{sec:Detec}
\subsubsection{Implementation Details} 

\noindent\textbf{Network structures.}
We
experiment with 
the recent one-stage architecture FCOS~\cite{tian2019fcos},
using  the backbone  ResNeXt-$32$x$8$d-$101$-FPN as the  teacher  network. The channel in the detector towers is set to $256$. It is a simple anchor-free model, but can achieve comparable performance with state-of-the-art two-stage  detection methods.

We choose two different models based on the MobileNetV$2$ backbone: c$128$-MNV$2$ and c$256$-MNV$2$ released by FCOS~\cite{tian2019fcos} as our student nets, where $c$ represents the channel in the detector towers. We apply the distillation loss on all the output levels of the feature pyramid network.

\noindent\textbf{Training setup.}
We follow the training schedule in FCOS~\cite{tian2019fcos}. For ablation studies,  all the teacher, the student w/ and w/o distillation are trained with stochastic gradient descent (SGD) for $90$K iterations with the initial learning rate being $0.01$ and a mini batch of $16$ images. The learning rate is reduced by a factor of $10$ at iteration $60$K and $80$K, respectively. Weight decay and momentum are set 
to be 
$0.0001$ and $0.9$, respectively. To compare with other state-of-the-art real-time detectors, we double the training iterations and the batch size, and the distillation method can further improve the results on the strong baselines.

\subsubsection{Dataset} 
\noindent\textbf{COCO.}
Microsoft Common Objects in Context (COCO)~\cite{lin2014microsoft} is a large-scale detection benchmark in object detection. 
There are $115K$ images for training and $5K$ images for validation. We evaluate the ablation results on the validation set, and we also submit the final results to the test-dev of COCO.

\subsubsection{Evaluation Metrics}
Average precision (AP) computes the average precision value for recall value over $0$ to $1$. The mAP
is the averaged AP over multiple Intersection over Union (IoU) values, from $0.5$ to $0.95$ with a step of $0.05$. 
We also report AP$50$ and AP$75$ represents for the AP with a single IoU of $0.5$ and $0.75$, respectively. APs, APm and APl are AP across different scales for small, medium and large objects.
\begin{table}[]
\centering
\caption{\textbf{Object detection.} PA vs.\  MIMIC on the COCO-minival split with MobileNetV$2$-c$256$ as the student net. Both distillation method can improve the accuracy of the detector, and the structured knowledge distillation performs better than the pixel-wise MIMIC. By applying all the distillation terms, the results can be further improved.}
\begin{tabular}{c|cccccc}
\hline
Method  & mAP  & AP50 & AP75 & APs  & APm  & APl  \\ \hline
Teacher & $42.5$ &  $61.7$      &   $45.9$     &    $26.0$  &  $46.2$    &   $54.3$   \\ \hline
student & $31.0$ & $48.5$   & $32.7$   & $17.1$ & $34.2$ & $39.7$ \\ \hline
+MIMIC~\cite{Li2017MimickingVE}  & $31.4$ & $48.1$   & $33.4$   & $16.5$ & $34.3$ & $41.0$ \\ \hline
+PA     & $31.9$ & $49.2$   & $33.7$  &$ 17.7$ & $34.7$ &$\bf{ 41.3}$ \\ \hline
Ours    & $\bf{32.1}$ & $\bf{49.5}$   &$\bf{ 34.2}$   & $\bf{18.5}$ & $\bf{35.3}$ & $41.2$ \\ \hline
\end{tabular}
\vspace{0.1cm}

\label{tab:oba}
\end{table}

\subsubsection{Results}

\noindent\textbf{Comparison with different distillation methods.}
To demonstrate the effectiveness of the structured knowledge distillation, we compare the pair-wise distillation method with the previous MIMIC~\cite{Li2017MimickingVE} method, which aligns the feature map on pixel-level. We use the c$256$-MNV$2$ as the student net and the results are shown in Table~\ref{tab:oba}.
By adding the pixel-wise MIMIC distillation method, the detector can be improved  by 
$0.4\%$ 
in 
mAP.
Our structured knowledge distillation method can improve by $0.9\%$ 
in
mAP. Under all evaluation metrics, the structured knowledge distillation method performs better than MIMIC. By combing the structured knowledge distillation with the pixel-wise distillation, the results can be further improved to $32.1\%$ mAP.
Comparing  to the baseline method without distillation, the improvement of AP$75$, APs and APl are more %
sound, 
indicating  the effectiveness of the distillation method.

\begin{figure*}[h]

\centering

\subfloat[Detection results w/o distillation]{

\begin{minipage}[b]{0.93\textwidth}

\includegraphics[width=0.94\textwidth]{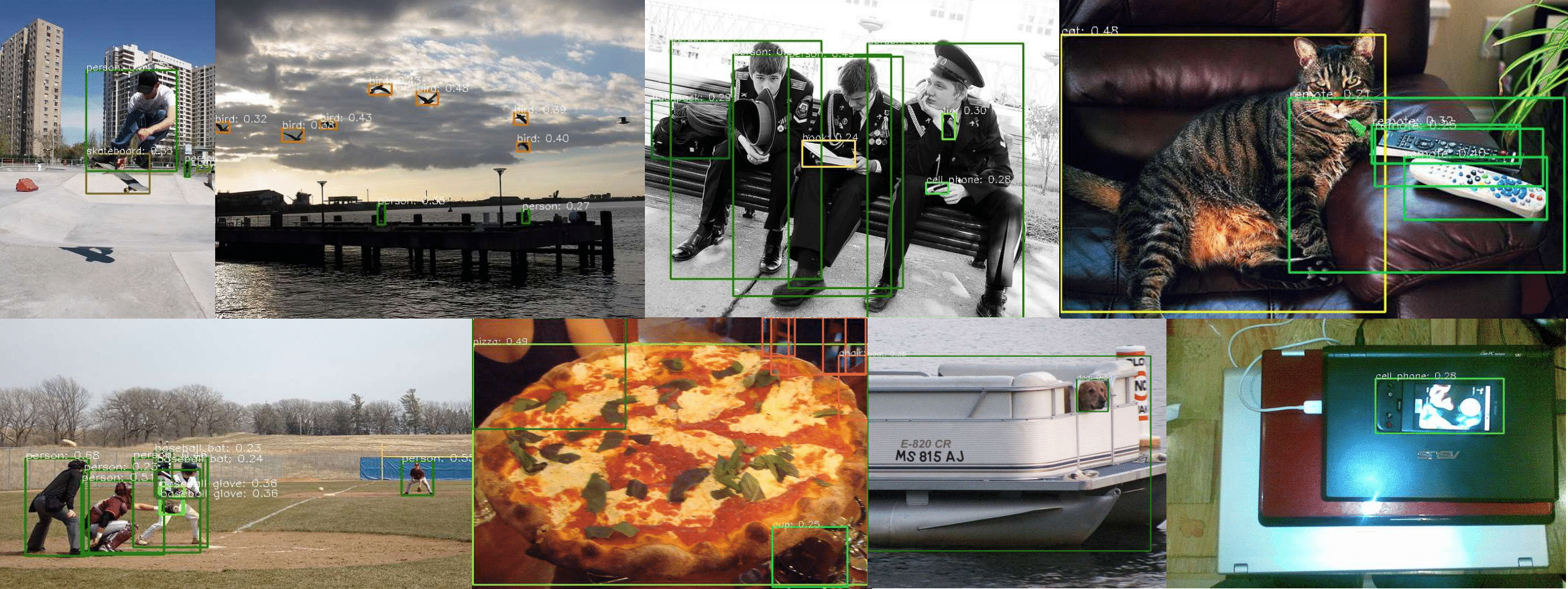}

\end{minipage}

}

\subfloat[Detection results w/ distillation]{

\begin{minipage}[b]{0.93\textwidth}

\includegraphics[width=0.93\textwidth]{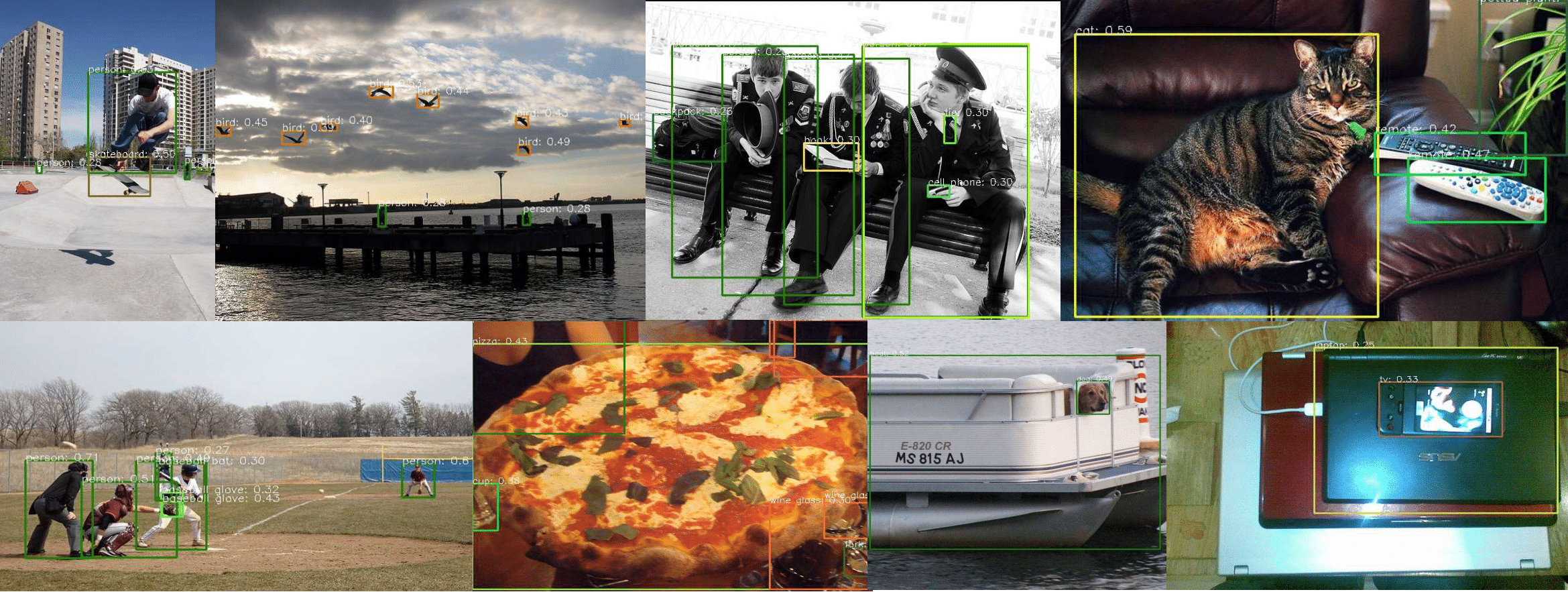}

\end{minipage}

}

 \caption{Detection results on the COCO dataset. With the structured knowledge distillation, the detector's accuracy is improved,  particularly for detecting 
 occluded, highly overlapped and extremely small objects. 
 } \label{fig:det}

\end{figure*}

We 
show some detection results 
in Figure~\ref{fig:det}. One can see that the detector trained with our 
distillation method can 
detect 
more small objects %
such as 
`person' and `bird'.

\begin{table}[t]
\centering
\caption{Detection accuracy with and without distillation on COCO-minival.}
\begin{tabular}{c|cccccc}
\hline
Method  & mAP  & AP$50$ & AP$75$ & APs  & APm  & APl  \\ \hline
Teacher & $42.5$ &  $61.7$      &   $45.9$     &    $26.0$  &  $46.2$    &  $ 54.3$   \\ \hline

C$128$-MV$2$ & $30.9$ & $48.5$   & $32.7$   & $17.1$ & $34.2$ & $39.7$ \\ \hline
w/ distillation  & $31.8$   &$49.2$    &$33.8$  &$17.8$  &$35.0$& $40.4$  \\ \hline\hline

C$256$-MV$2$ & $33.1$ & $51.1$   & $35.0$  & $18.5$ & $36.1$ & $43.4$ \\ \hline
w/ distillation& $33.9$ &  $51.8$  & $35.7$   & $19.7$ & $37.3$ & $43.4$ \\ \hline
\end{tabular}

\label{tab:differ}
\end{table}

\noindent\textbf{Results of different student nets.}
We follow the same training steps ($90$K) and batch size ($32$) as in FCOS~\cite{tian2019fcos} and apply the distillation method on two different
structures: C$256$-MV$2$ and C$128$-MV$2$. The results of w/ and w/o distillation are shown in Table~\ref{tab:differ}. 
By applying the structured knowledge distillation combine with pixel-wise distillation, the mAP of C$128$-MV$2$ and and C$256$-MV$2$ are improved by $0.9$ and $0.8$, respectively.

\begin{table*}[t]
\centering
\caption{Detection results and inference time on the COCO test-dev. The inference time %
was 
reported in the original papers~\cite{tian2019fcos,lin2017focal}. 
Our distillation method can improve the accuracy of a strong baseline with no extra inference time.}
\begin{tabular}{l|c|ccc|ccc|c}
 & backbone
 & AP & AP$_{50}$ & AP$_{75}$
 & AP$_S$ & AP$_M$ &  AP$_L$&time (ms/img)\\ [.1em]
\hline
 ~RetinaNet \cite{lin2017focal} & ResNet-$101$-FPN
  & $39.1$ & $59.1$ & $42.3$ & $21.8$ & $42.7$ & $50.2$ &$198$\\
 ~RetinaNet \cite{lin2017focal} & ResNeXt-$101$-FPN
  & $40.8$ & $61.1$ & $44.1$ & $24.1$ & $44.2$ & $51.2$ &-\\
  ~FCOS~\cite{tian2019fcos} (teacher) & ResNeXt-$101$-FPN&
$\bf{42.7}$& $\bf{62.2}$& $\bf{46.1}$ &$\bf{26.0}$& $\bf{45.6}$ &$\bf{52.6}$&$\bf{130}$\\
  \hline
  \hline
 ~YOLOv$2$ \cite{redmon2017yolo9000} & DarkNet-$19$
  & $21.6$ & $44.0$ & $19.2$ & $5.0$ &$22.4$ & $35.5$&$\bf{25}$ \\
 ~SSD$513$ \cite{liu2016ssd} & ResNet-$101$-SSD
  & $31.2$ & $50.4$ & $33.3$ & $10.2$ & $34.5$ & $49.8$&$125$ \\
 ~DSSD$513$ \cite{fu2017dssd} & ResNet-$101$-DSSD
  & $33.2$ & $53.3$ & $35.2$ & $13.0$ & $35.4$ & $51.1$ &$156$\\
  ~YOLOv$3$  \cite{redmon2018yolov3} & Darknet-$53$
  & $33.0$ & $\bf{57.9}$ & $34.4$ & $18.3$ & $35.4$ & $41.9$&$51$ \\
  ~FCOS (student)~\cite{tian2019fcos} & MobileNetV$2$-FPN
  & $31.4$ &$49.2$  &$33.3$  &$17.1$  &$33.5$  &$38.8$ & $45$\\
  ~FCOS (student) w/ distillation & MobileNetV$2$-FPN
  & $\bf{34.1}$ & $52.2$ &$\bf{ 36.4}$ & $\bf{19.0}$ &$\bf{36.2}$ & $\bf{42.0}$&$45$ \\
\end{tabular}

\label{results}
\end{table*}

\noindent\textbf{Results on the test-dev.}
The original mAP on the validation set of C$128$-MV$2$ reported by FCOS is $30.9\%$ with $90$K iterations. We double the training iterations and train with the distillation method. The final mAP on minival is $33.9\%$. 
The test results are in Table~\ref{results}, and we also list the AP and inference time for some state-of-the-art one-stage detectors to show the position of the baseline
and our detectors trained with the structured knowledge distillation method.
To make a fair comparison, we also double the training iterations without any distillation methods, and obtain mAP of $32.7\%$ on minival.

\section{Conclusion}
We have studied knowledge distillation
for training compact  
dense prediction networks
with the help of cumbersome/teacher networks.
By considering the structure information in dense prediction,
we have presented two structural distillation schemes: 
pair-wise distillation and holistic distillation.
We demonstrate the effectiveness of our proposed
distillation schemes on several recent compact networks
on three dense prediction tasks: semantic segmentation, depth estimation and object detection. Our structured knowledge distillation methods are complimentary to traditional pixel-wise distillation methods.

\section*{Acknowledgements}
C. Shen's participation was in part supported by ARC DP Project ``Deep learning that scales".

{\small
\bibliographystyle{IEEEtran}
\bibliography{main}

\begin{thebibliography}{10}
\providecommand{\url}[1]{#1}
\csname url@samestyle\endcsname
\providecommand{\newblock}{\relax}
\providecommand{\bibinfo}[2]{#2}
\providecommand{\BIBentrySTDinterwordspacing}{\spaceskip=0pt\relax}
\providecommand{\BIBentryALTinterwordstretchfactor}{4}
\providecommand{\BIBentryALTinterwordspacing}{\spaceskip=\fontdimen2\font plus
\BIBentryALTinterwordstretchfactor\fontdimen3\font minus
  \fontdimen4\font\relax}
\providecommand{\BIBforeignlanguage}[2]{{%
\expandafter\ifx\csname l@#1\endcsname\relax
\typeout{** WARNING: IEEEtran.bst: No hyphenation pattern has been}%
\typeout{** loaded for the language `#1'. Using the pattern for}%
\typeout{** the default language instead.}%
\else
\language=\csname l@#1\endcsname
\fi
#2}}
\providecommand{\BIBdecl}{\relax}
\BIBdecl

\bibitem{shelhamer2017fully}
E.~Shelhamer, J.~Long, and T.~Darrell, ``Fully convolutional networks for
  semantic segmentation.'' \emph{{IEEE} Trans. Pattern Anal. Mach. Intell.},
  vol.~39, no.~4, p. 640, 2017.

\bibitem{chen2018deeplab}
L.-C. Chen, G.~Papandreou, I.~Kokkinos, K.~Murphy, and A.~L. Yuille, ``Deeplab:
  Semantic image segmentation with deep convolutional nets, atrous convolution,
  and fully connected crfs,'' \emph{{IEEE} Trans. Pattern Anal. Mach. Intell.},
  vol.~40, no.~4, pp. 834--848, 2018.

\bibitem{zhao2017pyramid}
H.~Zhao, J.~Shi, X.~Qi, X.~Wang, and J.~Jia, ``Pyramid scene parsing network,''
  in \emph{Proc. IEEE Conf. Comp. Vis. Patt. Recogn.}, 2017, pp. 2881--2890.

\bibitem{lin2019refinenet}
G.~Lin, F.~Liu, A.~Milan, C.~Shen, and I.~Reid, ``Refinenet: Multi-path
  refinement networks for dense prediction,'' \emph{{IEEE} Trans. Pattern Anal.
  Mach. Intell.}, 2019.

\bibitem{tian2019fcos}
Z.~Tian, C.~Shen, H.~Chen, and T.~He, ``{FCOS}: Fully convolutional one-stage
  object detection,'' \emph{Proc. IEEE Int. Conf. Comp. Vis.}, 2019.

\bibitem{paszke2016enet}
A.~Paszke, A.~Chaurasia, S.~Kim, and E.~Culurciello, ``Enet: A deep neural
  network architecture for real-time semantic segmentation,'' \emph{arXiv:
  Comp. Res. Repository}, vol. abs/1606.02147, 2016.

\bibitem{zhao2017icnet}
H.~Zhao, X.~Qi, X.~Shen, J.~Shi, and J.~Jia, ``Icnet for real-time semantic
  segmentation on high-resolution images,'' \emph{Proc. Eur. Conf. Comp. Vis.},
  2018.

\bibitem{redmon2016you}
J.~Redmon, S.~Divvala, R.~Girshick, and A.~Farhadi, ``You only look once:
  Unified, real-time object detection,'' in \emph{Proc. IEEE Conf. Comp. Vis.
  Patt. Recogn.}, 2016, pp. 779--788.

\bibitem{liu2016ssd}
W.~Liu, D.~Anguelov, D.~Erhan, C.~Szegedy, S.~Reed, C.-Y. Fu, and A.~C. Berg,
  ``Ssd: Single shot multibox detector,'' in \emph{Proc. Eur. Conf. Comp.
  Vis.}\hskip 1em plus 0.5em minus 0.4em\relax Springer, 2016, pp. 21--37.

\bibitem{wofk2019fastdepth}
D.~Wofk, F.~Ma, T.-J. Yang, S.~Karaman, and V.~Sze, ``Fastdepth: Fast monocular
  depth estimation on embedded systems,'' \emph{Int. Conf. on Robotics and
  Automation}, 2019.

\bibitem{Li2017MimickingVE}
Q.~Li, S.~Jin, and J.~Yan, ``Mimicking very efficient network for object
  detection,'' \emph{Proc. IEEE Conf. Comp. Vis. Patt. Recogn.}, pp.
  7341--7349, 2017.

\bibitem{xieimproving}
J.~Xie, B.~Shuai, J.-F. Hu, J.~Lin, and W.-S. Zheng, ``Improving fast
  segmentation with teacher-student learning,'' \emph{Proc. British Machine
  Vis. Conf.}, 2018.

\bibitem{badrinarayanan2017segnet}
V.~Badrinarayanan, A.~Kendall, and R.~Cipolla, ``Segnet: A deep convolutional
  encoder-decoder architecture for image segmentation,'' \emph{{IEEE} Trans.
  Pattern Anal. Mach. Intell.}, no.~12, pp. 2481--2495, 2017.

\bibitem{romera2017efficient}
E.~Romera, J.~M. Alvarez, L.~M. Bergasa, and R.~Arroyo, ``Efficient convnet for
  real-time semantic segmentation,'' in \emph{IEEE Intelligent Vehicles Symp.},
  2017, pp. 1789--1794.

\bibitem{mehta2018espnet}
S.~Mehta, M.~Rastegari, A.~Caspi, L.~Shapiro, and H.~Hajishirzi, ``Espnet:
  Efficient spatial pyramid of dilated convolutions for semantic
  segmentation,'' \emph{Proc. Eur. Conf. Comp. Vis.}, 2018.

\bibitem{LightNet}
H.~Liu, ``Lightnet: Light-weight networks for semantic image segmentation,''
  \url{https://github.com/ansleliu/LightNet}, 2018.

\bibitem{Yuan2018OCNetOC}
Y.~Yuan and J.~Wang, ``Ocnet: Object context network for scene parsing,'' in
  \emph{arXiv: Comp. Res. Repository}, vol. abs/1809.00916, 2018.

\bibitem{hinton2015distilling}
G.~E. Hinton, O.~Vinyals, and J.~Dean, ``Distilling the knowledge in a neural
  network,'' \emph{arXiv: Comp. Res. Repository}, vol. abs/1503.02531, 2015.

\bibitem{romero2014fitnets}
A.~Romero, N.~Ballas, S.~E. Kahou, A.~Chassang, C.~Gatta, and Y.~Bengio,
  ``Fitnets: Hints for thin deep nets,'' \emph{arXiv: Comp. Res. Repository},
  vol. abs/1412.6550, 2014.

\bibitem{li2009markov}
S.~Z. Li, \emph{Markov random field modeling in image analysis}.\hskip 1em plus
  0.5em minus 0.4em\relax Springer Science \& Business Media, 2009.

\bibitem{He2016DeepRL}
K.~He, X.~Zhang, S.~Ren, and J.~Sun, ``Deep residual learning for image
  recognition,'' \emph{Proc. IEEE Conf. Comp. Vis. Patt. Recogn.}, pp.
  770--778, 2016.

\bibitem{huang2017densely}
G.~Huang, Z.~Liu, L.~van~der Maaten, and K.~Q. Weinberger, ``Densely connected
  convolutional networks,'' \emph{Proc. IEEE Conf. Comp. Vis. Patt. Recogn.},
  pp. 2261--2269, 2017.

\bibitem{yu2015multi}
F.~Yu and V.~Koltun, ``Multi-scale context aggregation by dilated
  convolutions,'' \emph{Proc. Int. Conf. Learn. Representations}, 2016.

\bibitem{lin2016efficient}
G.~Lin, C.~Shen, A.~van~den Hengel, and I.~Reid, ``Efficient piecewise training
  of deep structured models for semantic segmentation,'' in \emph{Proc. IEEE
  Conf. Comp. Vis. Patt. Recogn.}, 2016, pp. 3194--3203.

\bibitem{SzegedyVISW16}
C.~Szegedy, V.~Vanhoucke, S.~Ioffe, J.~Shlens, and Z.~Wojna, ``Rethinking the
  inception architecture for computer vision,'' \emph{Proc. IEEE Conf. Comp.
  Vis. Patt. Recogn.}, pp. 2818--2826, 2016.

\bibitem{treml2016speeding}
M.~Treml, J.~Arjona-Medina, T.~Unterthiner, R.~Durgesh, F.~Friedmann,
  P.~Schuberth, A.~Mayr, M.~Heusel, M.~Hofmarcher, M.~Widrich \emph{et~al.},
  ``Speeding up semantic segmentation for autonomous driving,'' in \emph{Proc.
  Workshop of Advances in Neural Inf. Process. Syst.}, 2016.

\bibitem{IandolaMAHDK16}
F.~N. Iandola, M.~W. Moskewicz, K.~Ashraf, S.~Han, W.~J. Dally, and K.~Keutzer,
  ``Squeezenet: Alexnet-level accuracy with 50x fewer parameters and
  {\textless}1mb model size,'' \emph{arXiv: Comp. Res. Repository}, vol.
  abs/1602.07360, 2016.

\bibitem{howard2017mobilenets}
A.~G. Howard, M.~Zhu, B.~Chen, D.~Kalenichenko, W.~Wang, T.~Weyand,
  M.~Andreetto, and H.~Adam, ``Mobilenets: Efficient convolutional neural
  networks for mobile vision applications,'' \emph{arXiv: Comp. Res.
  Repository}, vol. abs/1704.04861, 2017.

\bibitem{Zhang2017ShuffleNet}
X.~Zhang, X.~Zhou, M.~Lin, and J.~Sun, ``Shufflenet: An extremely efficient
  convolutional neural network for mobile devices,'' \emph{Proc. IEEE Conf.
  Comp. Vis. Patt. Recogn.}, 2018.

\bibitem{saxena2007learning}
A.~Saxena, M.~Sun, and A.~Y. Ng, ``Learning 3-d scene structure from a single
  still image,'' in \emph{Proc. IEEE Int. Conf. Comp. Vis.}\hskip 1em plus
  0.5em minus 0.4em\relax IEEE, 2007, pp. 1--8.

\bibitem{eigen}
D.~Eigen, C.~Puhrsch, and R.~Fergus, ``Depth map prediction from a single image
  using a multi-scale deep network,'' in \emph{Proc. Advances in Neural Inf.
  Process. Syst.}, 2014.

\bibitem{CVPR15Liu}
F.~Liu, C.~Shen, and G.~Lin, ``Deep convolutional neural fields for depth
  estimation from a single image,'' in \emph{Proc. IEEE Conf. Comp. Vis. Patt.
  Recogn.}, 2015.

\bibitem{Depth2015Liu}
F.~Liu, C.~Shen, G.~Lin, and I.~Reid, ``Learning depth from single monocular
  images using deep convolutional neural fields,'' \emph{{IEEE} Trans. Pattern
  Anal. Mach. Intell.}, 2016.

\bibitem{li2018deep}
R.~Li, K.~Xian, C.~Shen, Z.~Cao, H.~Lu, and L.~Hang, ``Deep attention-based
  classification network for robust depth prediction,'' in \emph{Proc. Asian
  Conf. Comp. Vis.}, 2018.

\bibitem{fu2018deep}
H.~Fu, M.~Gong, C.~Wang, K.~Batmanghelich, and D.~Tao, ``Deep ordinal
  regression network for monocular depth estimation,'' in \emph{Proc. IEEE
  Conf. Comp. Vis. Patt. Recogn.}, 2018, pp. 2002--2011.

\bibitem{fei2018geo}
X.~Fei, A.~Wang, and S.~Soatto, ``Geo-supervised visual depth prediction,'' in
  \emph{arXiv: Comp. Res. Repository}, vol. abs/1807.11130, 2018.

\bibitem{wei2019enforcing}
Y.~Wei, Y.~Liu, C.~Shen, and Y.~Yan, ``Enforcing geometric constraints of
  virtual normal for depth prediction,'' \emph{Proc. IEEE Int. Conf. Comp.
  Vis.}, 2019.

\bibitem{spek2018cream}
A.~Spek, T.~Dharmasiri, and T.~Drummond, ``Cream: Condensed real-time models
  for depth prediction using convolutional neural networks,'' in \emph{Int.
  Conf. on Intell. Robots and Sys.}\hskip 1em plus 0.5em minus 0.4em\relax
  IEEE, 2018, pp. 540--547.

\bibitem{girshick2015fast}
R.~Girshick, ``Fast r-cnn,'' in \emph{Proc. IEEE Int. Conf. Comp. Vis.}, 2015,
  pp. 1440--1448.

\bibitem{ren2015faster}
S.~Ren, K.~He, R.~Girshick, and J.~Sun, ``Faster r-cnn: Towards real-time
  object detection with region proposal networks,'' in \emph{Proc. Advances in
  Neural Inf. Process. Syst.}, 2015, pp. 91--99.

\bibitem{lin2017focal}
T.-Y. Lin, P.~Goyal, R.~Girshick, K.~He, and P.~Doll{\'a}r, ``Focal loss for
  dense object detection,'' in \emph{Proc. IEEE Int. Conf. Comp. Vis.}, 2017,
  pp. 2980--2988.

\bibitem{ba2014deep}
J.~Ba and R.~Caruana, ``Do deep nets really need to be deep?'' in \emph{Proc.
  Advances in Neural Inf. Process. Syst.}, 2014, pp. 2654--2662.

\bibitem{Urban2016Do}
G.~Urban, K.~J. Geras, S.~E. Kahou, O.~Aslan, S.~Wang, R.~Caruana, A.~Mohamed,
  M.~Philipose, and M.~Richardson, ``Do deep convolutional nets really need to
  be deep (or even convolutional)?'' in \emph{Proc. Int. Conf. Learn.
  Representations}, 2016.

\bibitem{Zagoruyko2016PayingMA}
S.~Zagoruyko and N.~Komodakis, ``Paying more attention to attention: Improving
  the performance of convolutional neural networks via attention transfer,''
  \emph{Proc. Int. Conf. Learn. Representations}, 2017.

\bibitem{liu2019structured}
Y.~Liu, K.~Chen, C.~Liu, Z.~Qin, Z.~Luo, and J.~Wang, ``Structured knowledge
  distillation for semantic segmentation,'' in \emph{Proc. IEEE Conf. Comp.
  Vis. Patt. Recogn.}, 2019, pp. 2604--2613.

\bibitem{Sandler2018MobileNetV2IR}
M.~Sandler, A.~Howard, M.~Zhu, A.~Zhmoginov, and L.-C. Chen, ``Mobilenetv2:
  Inverted residuals and linear bottlenecks,'' in \emph{Proc. IEEE Conf. Comp.
  Vis. Patt. Recogn.}, 2018.

\bibitem{wang2018text}
H.~Wang, Z.~Qin, and T.~Wan, ``Text generation based on generative adversarial
  nets with latent variables,'' in \emph{Proc. Pacific-Asia Conf. Knowledge
  discovery \& data mining}, 2018, pp. 92--103.

\bibitem{yu2017seqgan}
L.~Yu, W.~Zhang, J.~Wang, and Y.~Yu, ``Seqgan: Sequence generative adversarial
  nets with policy gradient.'' in \emph{Proc. {AAAI} Conf. Artificial Intell.},
  2017, pp. 2852--2858.

\bibitem{Goodfellow2014Generative}
I.~J. Goodfellow, J.~Pougetabadie, M.~Mirza, B.~Xu, D.~Wardefarley, S.~Ozair,
  A.~Courville, Y.~Bengio, Z.~Ghahramani, and M.~Welling, ``Generative
  adversarial nets,'' \emph{Proc. Advances in Neural Inf. Process. Syst.},
  vol.~3, pp. 2672--2680, 2014.

\bibitem{karras2017progressive}
T.~Karras, T.~Aila, S.~Laine, and J.~Lehtinen, ``Progressive growing of gans
  for improved quality, stability, and variation,'' \emph{Proc. Int. Conf.
  Learn. Representations}, 2018.

\bibitem{MirzaO14}
M.~Mirza and S.~Osindero, ``Conditional generative adversarial nets,''
  \emph{arXiv: Comp. Res. Repository}, vol. abs/1411.1784, 2014.

\bibitem{Johnson2016Perceptual}
J.~Johnson, A.~Alahi, and L.~Fei-Fei, ``Perceptual losses for real-time style
  transfer and super-resolution,'' \emph{Proc. Eur. Conf. Comp. Vis.}, pp.
  694--711, 2016.

\bibitem{Yifan2018Auto}
Y.~Liu, Z.~Qin, T.~Wan, and Z.~Luo, ``Auto-painter: Cartoon image generation
  from sketch by using conditional wasserstein generative adversarial
  networks,'' \emph{Neurocomputing}, vol. 311, pp. 78--87, 2018.

\bibitem{chen2017adversarial}
Y.~Chen, C.~Shen, X.-S. Wei, L.~Liu, and J.~Yang, ``Adversarial {PoseNet}: A
  structure-aware convolutional network for human pose estimation,'' in
  \emph{Proc. IEEE Int. Conf. Comp. Vis.}, 2017, pp. 1212--1221.

\bibitem{luc2016semantic}
P.~Luc, C.~Couprie, S.~Chintala, and J.~Verbeek, ``Semantic segmentation using
  adversarial networks,'' \emph{arXiv: Comp. Res. Repository}, vol.
  abs/1611.08408, 2016.

\bibitem{gwn2018generative}
K.~Gwn~Lore, K.~Reddy, M.~Giering, and E.~A. Bernal, ``Generative adversarial
  networks for depth map estimation from rgb video,'' in \emph{Proc. IEEE Conf.
  Comp. Vis. Patt. Recogn.}, 2018, pp. 1177--1185.

\bibitem{gulrajani2017improved}
I.~Gulrajani, F.~Ahmed, M.~Arjovsky, V.~Dumoulin, and A.~C. Courville,
  ``Improved training of wasserstein gans,'' in \emph{Proc. Advances in Neural
  Inf. Process. Syst.}, 2017, pp. 5767--5777.

\bibitem{Zhang2018Self}
H.~Zhang, I.~Goodfellow, D.~Metaxas, and A.~Odena, ``Self-attention generative
  adversarial networks,'' in \emph{arXiv: Comp. Res. Repository}, vol.
  abs/1805.08318, 2018.

\bibitem{cao2017estimating}
Y.~Cao, Z.~Wu, and C.~Shen, ``Estimating depth from monocular images as
  classification using deep fully convolutional residual networks,''
  \emph{{IEEE} Trans. Circuits Syst. Video Technol.}, vol.~28, no.~11, pp.
  3174--3182, 2017.

\bibitem{Cordts2016Cityscapes}
M.~Cordts, M.~Omran, S.~Ramos, T.~Rehfeld, M.~Enzweiler, R.~Benenson,
  U.~Franke, S.~Roth, and B.~Schiele, ``The cityscapes dataset for semantic
  urban scene understanding,'' in \emph{Proc. IEEE Conf. Comp. Vis. Patt.
  Recogn.}, 2016.

\bibitem{brostow2008segmentation}
G.~J. Brostow, J.~Shotton, J.~Fauqueur, and R.~Cipolla, ``Segmentation and
  recognition using structure from motion point clouds,'' in \emph{Proc. Eur.
  Conf. Comp. Vis.}\hskip 1em plus 0.5em minus 0.4em\relax Springer, 2008, pp.
  44--57.

\bibitem{zhou2017scene}
B.~Zhou, H.~Zhao, X.~Puig, S.~Fidler, A.~Barriuso, and A.~Torralba, ``Scene
  parsing through ade20k dataset,'' in \emph{Proc. IEEE Conf. Comp. Vis. Patt.
  Recogn.}, 2017.

\bibitem{FLOPs}
\url{https://github.com/warmspringwinds/pytorch-segmentation-detection/blob/master/pytorch_segmentation_detection/utils/flops_benchmark.py},
  2018.

\bibitem{drozdzalone}
S.~J.~M. Drozdzal, D.~Vazquez, and A.~R.~Y. Bengio, ``The one hundred layers
  tiramisu: Fully convolutional densenets for semantic segmentation,''
  \emph{Proc. Workshop of IEEE Conf. Comp. Vis. Patt. Recogn.}, 2017.

\bibitem{chen2014semantic}
L.-C. Chen, G.~Papandreou, I.~Kokkinos, K.~Murphy, and A.~Yuille, ``Semantic
  image segmentation with deep convolutional nets and fully connected crfs,''
  in \emph{Proc. Int. Conf. Learn. Representations}, 2015.

\bibitem{xiao2018unified}
T.~Xiao, Y.~Liu, B.~Zhou, Y.~Jiang, and J.~Sun, ``Unified perceptual parsing
  for scene understanding,'' in \emph{Proc. Eur. Conf. Comp. Vis.}, 2018.

\bibitem{laina2016deeper}
I.~Laina, C.~Rupprecht, V.~Belagiannis, F.~Tombari, and N.~Navab, ``Deeper
  depth prediction with fully convolutional residual networks,'' in
  \emph{Proc.\ Int.\ Conf.\ 3D Vision (3DV)}.\hskip 1em plus 0.5em minus
  0.4em\relax IEEE, 2016, pp. 239--248.

\bibitem{hu2019revisiting}
J.~Hu, M.~Ozay, Y.~Zhang, and T.~Okatani, ``Revisiting single image depth
  estimation: Toward higher resolution maps with accurate object boundaries,''
  in \emph{Proc. Winter Conf. on Appl. of Comp0 Vis.}\hskip 1em plus 0.5em
  minus 0.4em\relax IEEE, 2019, pp. 1043--1051.

\bibitem{nekrasov2018real}
V.~Nekrasov, T.~Dharmasiri, A.~Spek, T.~Drummond, C.~Shen, and I.~Reid,
  ``Real-time joint semantic segmentation and depth estimation using asymmetric
  annotations,'' in \emph{arXiv: Comp. Res. Repository}, vol. abs/1809.04766,
  2018.

\bibitem{lin2014microsoft}
T.-Y. Lin, M.~Maire, S.~Belongie, J.~Hays, P.~Perona, D.~Ramanan,
  P.~Doll{\'a}r, and C.~L. Zitnick, ``Microsoft coco: Common objects in
  context,'' in \emph{Proc. Eur. Conf. Comp. Vis.}\hskip 1em plus 0.5em minus
  0.4em\relax Springer, 2014, pp. 740--755.

\bibitem{redmon2017yolo9000}
J.~Redmon and A.~Farhadi, ``Yolo9000: better, faster, stronger,'' in
  \emph{Proc. IEEE Conf. Comp. Vis. Patt. Recogn.}, 2017, pp. 7263--7271.

\bibitem{fu2017dssd}
C.-Y. Fu, W.~Liu, A.~Ranga, A.~Tyagi, and A.~C. Berg, ``Dssd: Deconvolutional
  single shot detector,'' in \emph{arXiv: Comp. Res. Repository}, vol.
  abs/1701.06659, 2017.

\bibitem{redmon2018yolov3}
J.~Redmon and A.~Farhadi, ``Yolov3: An incremental improvement,'' in
  \emph{arXiv: Comp. Res. Repository}, vol. abs/1804.02767, 2018.

\end{thebibliography}
}
\vspace{-10 mm}
\begin{IEEEbiography}[{\includegraphics[width=1in,height=1.25in,clip,keepaspectratio]{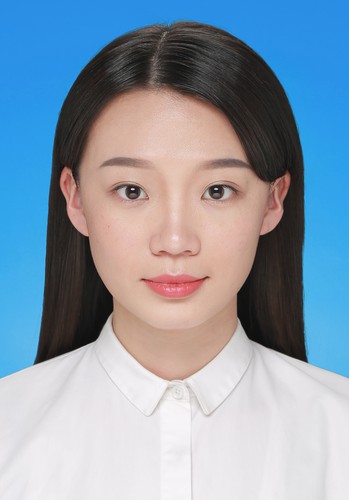}}]{Yifan Liu} is 
a Ph.D candidate in Computer Science at The University of Adelaide, supervised by Professor Chunhua Shen. She obtained her B.S. and M.Sc. in Artificial Intelligence from Beihang University. Her research interests include image processing, dense prediction and real-time applications in deep learning.
\end{IEEEbiography}
\vspace{-13 mm}
\begin{IEEEbiography}
[{\includegraphics[width=1in,height=1.25in,clip,keepaspectratio]{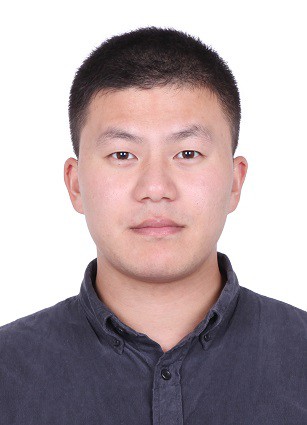}}]
{Changyong Shu}
received the Ph.D. degree 
from Beihang University, Beijing, China, in 2017. He has been with the 
Nanjing Institute of Advanced Artificial Intelligence since 2018.
His current research interest focuses on knowledge distillation.
 \end{IEEEbiography}
 \vspace{-13 mm}
\begin{IEEEbiography}[{\includegraphics[width=1in, clip,keepaspectratio]{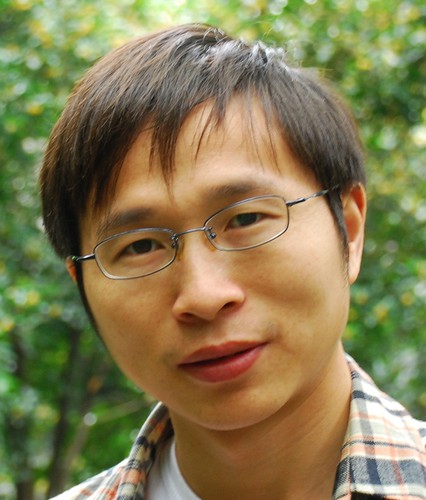}}]{Jingdong Wang}
is a Senior Principal Research Manager with the Visual Computing Group,
Microsoft Research, Beijing, China.
He received the B.Eng. and M.Eng. degrees
from the  Department of Automation, Tsinghua University, Beijing, China, in 2001 and 2004, respectively,
and the PhD degree
from the Department of Computer Science and Engineering,
the Hong Kong University of Science and Technology,
Hong Kong, in 2007.
His areas of interest include deep learning, large-scale indexing, human understanding, and person re-identification.
He is an Associate Editor of IEEE TPAMI,
IEEE TMM and IEEE TCSVT,
and is an area chair (or SPC) of some prestigious conferences,
such as CVPR, ICCV, ECCV, ACM MM, IJCAI, and AAAI.
He is a Fellow of IAPR and an ACM Distinguished Member.
\end{IEEEbiography}
\vspace{-13 mm}
\begin{IEEEbiographynophoto}
{Chunhua Shen}
 is a Professor at School of Computer Science, The University of Adelaide,
 Australia.
\end{IEEEbiographynophoto}

\clearpage
\newpage

\end{document}